\documentclass[12pt, a4paper]{article}
\usepackage{xr}
\usepackage{array,bm,amsmath,amssymb,mathrsfs,epsfig,epstopdf}
\usepackage{amsthm}

\usepackage{setspace} 
\usepackage{CJK}
\usepackage{color}
\usepackage{lscape}
\usepackage{verbatim}
\usepackage{enumerate}
\usepackage{booktabs}
\usepackage{threeparttable}
\usepackage{multicol} 
\usepackage{longtable}
\usepackage{multirow}
\usepackage{bm}
\usepackage{bbm}
\usepackage{cases}
\usepackage{caption}
\usepackage{natbib}
\usepackage{algorithm}
\usepackage{algorithmic}
\usepackage{geometry}
\usepackage{framed}
\usepackage[pdftex,bookmarksnumbered,bookmarksopen,colorlinks, citecolor=blue,linkcolor=blue,urlcolor=blue]{hyperref}
\usepackage{graphicx}
\usepackage{amsmath}
\usepackage{amsfonts}
\usepackage{float}
\usepackage{ragged2e}
\usepackage[figuresright]{rotating}
\usepackage{mathtools}
\usepackage{pifont}
\usepackage{indentfirst}  
\usepackage{listings}
\definecolor{codegreen}{rgb}{0,0.6,0}
\definecolor{codegray}{rgb}{0.5,0.5,0.5}
\definecolor{codepurple}{rgb}{0.58,0,0.82}
\definecolor{backcolour}{rgb}{0.95,0.95,0.92}
\lstdefinestyle{mystyle}{
	backgroundcolor=\color{backcolour},   
	commentstyle=\color{codegreen},
	keywordstyle=\color{magenta},
	numberstyle=\tiny\color{codegray},
	stringstyle=\color{codepurple},
	basicstyle=\ttfamily\footnotesize,
	breakatwhitespace=false,         
	breaklines=true,                 
	captionpos=b,                    
	keepspaces=true,                 
	numbers=left,                    
	numbersep=5pt,                  
	showspaces=false,                
	showstringspaces=false,
	showtabs=false,                  
	tabsize=2
}

\newcommand\Vector[1]{\bm{#1}}

\newcommand\va{{\Vector{a}}}

\newcommand\vg{{\Vector{g}}}

\newcommand\vs{{\Vector{s}}}

\newcommand\vw{{\Vector{w}}}
\newcommand\vx{{\Vector{x}}}

\newcommand\vbeta{{\Vector{\beta}}}
\newcommand\vgamma{{\Vector{\gamma}}}

\newcommand\vtheta{{\Vector{\theta}}}

\newcommand\vlambda{{\Vector{\lambda}}}
\newcommand\vmu{{\Vector{\mu}}}

\newcommand\MATRIX[1]{\bm{#1}}

\newcommand\mA{{\MATRIX{A}}}

\newcommand\mD{{\MATRIX{D}}}

\newcommand\mG{{\MATRIX{G}}}

\newcommand\mI{{\MATRIX{I}}}

\newcommand\mS{{\MATRIX{S}}}

\newcommand\mU{{\MATRIX{U}}}
\newcommand\mV{{\MATRIX{V}}}
\newcommand\mW{{\MATRIX{W}}}

\newcommand\mY{{\MATRIX{Y}}}
\newcommand\mZ{{\MATRIX{Z}}}

\newcommand\mGamma{{\MATRIX{\Gamma}}}

\newcommand\mTheta{{\MATRIX{\Theta}}}
\newcommand\mLambda{{\MATRIX{\Lambda}}}

\newcommand\mSigma{{\MATRIX{\Sigma}}}

\newcommand\gK{{\mathcal{K}}}

\newcommand\gT{{\mathcal{T}}}

\newcommand\gW{{\mathcal{W}}}

\DeclareMathOperator*{\argmin}{arg\,min}

\DeclareMathOperator{\sign}{sign}

\newtheorem{theorem}{Theorem}[section]
\newtheorem{lemma}{Lemma}[section]

\newcommand{\tabincell}[2]{\begin{tabular}{@{}#1@{}}#2\end{tabular}}

\title{\bf Block Toeplitz Sparse Precision Matrix Estimation for Large-Scale Interval-Valued Time Series Forecasting}
\author{Wan Tian$^1$, Zhongfeng Qin$^{1,2}$\thanks{Corresponding author}}
\date{
	wantian61@foxmail.com, qin@buaa.edu.cn\\
	$^1$ School of Economics and Management, Beihang University, Beijing 100191, China \\%
	$^2$Key Laboratory of Complex System Analysis, Management and Decision (Beihang University), Ministry of Education, Beijing 100191, China \\[2ex]%
}

\begin{document}
\maketitle

\begin{abstract}
Modeling and forecasting interval-valued time series (ITS) have attracted considerable attention due to their growing presence in various contexts. To the best of our knowledge, there have been no efforts to model large-scale ITS. In this paper, we propose a feature extraction procedure for large-scale ITS, which involves key steps such as auto-segmentation and clustering, and feature transfer learning. This procedure can be seamlessly integrated with any suitable prediction models for forecasting purposes. Specifically, we transform the automatic segmentation and clustering of ITS into the estimation of Toeplitz sparse precision matrices and assignment set. The majorization-minimization algorithm is employed to convert this highly non-convex optimization problem into two subproblems. We derive efficient dynamic programming and alternating direction method to solve these two subproblems alternately and establish their convergence properties. By employing the Joint Recurrence Plot (JRP) to image subsequence and assigning a class label to each cluster, an image dataset is constructed. Then, an appropriate neural network is chosen to train on this image dataset and used to extract features for the next step of forecasting. Real data applications demonstrate that the proposed method can effectively obtain invariant representations of the raw data and enhance forecasting performance.

\end{abstract}

\maketitle

\section{Introduction}
In recent years, there has been a growing interest in the modeling and analysis of interval-valued time series (ITS) within the realms of statistics and econometrics \citep{gonzalez2013constrained, han2016vector, sun2018threshold, sun2022model, sun2024nonparametric}. In general, the generation of ITS typically follows two primary approaches. One approach involves deliberately aggregating a substantial number of point-valued observations into intervals. This is aimed at reducing sample size and conserving storage space \citep{billard2003statistics}. The other approach relates to intervals that naturally emerge in practical applications. For instance, the daily maximum and minimum air quality indices in meteorological science, the maximum and minimum economic growth rates observed over a year in economics, and the highest and lowest prices of specific stocks throughout a trading day in financial markets, all constitute ITS. Modeling ITS offers two primary advantages over modeling point-valued time series. First, interval-valued observations contain more information about variation and level characteristics over the same period than point-valued observations \citep{gonzalez2013constrained, sun2022model}, leading to more efficient estimation and powerful inference. Second, certain types of disturbances may have a detrimental impact on inference from point-valued data, whereas these issues can be mitigated by modeling interval-valued time series in a low-frequency setup \citep{han2016vector}.

Extensive research in finance and econometrics has introduced various methods for modeling and forecasting univariate ITS. \citet{arroyo2007exponential} extended the exponential smoothing method to interval-valued scenarios by employing interval arithmetic, while simultaneously addressing seasonality and trend components. \citet{san2007imlp} introduced an interval multilayer perceptron model leveraging an interval neural network framework. The pivotal aspect of this model entails employing the monotonic nonlinear hyperbolic tangent as the activation function, thereby ensuring that the output of the neural network conforms to the relative magnitudes of the upper and lower bounds of the interval. \citet{maia2008forecasting} not only extended various classical models such as autoregressive, autoregressive integrated moving average, and multilayer perceptron to the interval-valued scenario, but also combined these models using a hybrid approach. There are also research efforts aimed at directly modeling the upper and lower bounds of intervals. For instance, \citet{gonzalez2013constrained} proposed a regression model for interval-valued time series with constraints, presenting two parameter estimation methods and demonstrating their consistency. The aforementioned studies primarily focus on modeling from the perspective of bivariate representation of intervals.

The aforementioned studies are based on traditional statistical learning methods for modeling bivariate point-valued series to achieve the goal of modeling the original ITS. Recently, some studies have attempted to model the interval as a whole. The pioneering work in this direction is by \citet{han2012autoregressive}. They ignored the size relationship between the upper and lower bounds of the intervals and introduced the concept of extended random intervals. Based on this concept and by selecting appropriate distance measures between paired intervals, they developed autoregressive conditional models for univariate ITS and the corresponding asymptotic theory. \citet{han2016vector} developed autoregressive moving average models for interval vector-valued time series and the corresponding parameter estimation theory. \citet{sun2018threshold} also built upon this concept, establishing the first nonlinear time series models for ITS, along with recent model averaging methods \citep{sun2022model} and non-parametric methods \citep{sun2024nonparametric}. Therefore, research on modeling multivariate ITS is still limited, not to mention the modeling and forecasting of large-scale ITS. However, in the financial domain, where ITS naturally occur most frequently, there is an urgent need for large-scale modeling and forecasting \citep{cao2023large}.

In recent years, deep learning technologies, with neural networks as their representative, have seen highly successful applications in various scenarios, sometimes even surpassing human performance \citep{DBLP:conf/iccv/HeZRS15}. Within the realm of point-valued time series modeling, recurrent neural networks (RNN) and their variants, such as gated recurrent units (GRU) and long short-term memory (LSTM), as well as the recently popular Transformer architecture \citep{zhou2021informer}, have achieved state-of-the-art performance across tasks including forecasting \citep{DBLP:conf/nips/GuenT19, bandara2020forecasting}, classification \citep{fulcher2014highly}, clustering \citep{DBLP:conf/ijcai/LeiYVWD19}, and anomaly detection \citep{corizzo2020scalable}.

Generally, the effectiveness of deep learning technologies stems from their ability to learn effective representations of raw data \citep{lecun2015deep}. 
Certainly, a considerable amount of research has been conducted to construct modeling methods for large-scale point-valued time series based on deep learning technologies. For instance, \citet{du2021adarnn} proposed AdaRNN, which comprises two modules aimed at describing, learning, and matching distributional features. \citet{cao2023large} introduced a regularizer based on Invariant Risk Minimization (IRM) \citep{arjovsky2019invariant} to address the issue of distributional shifts. The essence of these large-scale modeling methods is to learn invariant deep representations of the raw data. However, to our knowledge, there are few works that currently model ITS from the perspective of deep learning, let alone modeling and forecasting large-scale ITS.

In this paper, we propose a feature extraction procedure that can effectively learn invariant representations of large-scale ITS, with block Toeplitz sparse precision matrix estimation and multivariate time series imaging as the core steps. This procedure can be integrated with suitable prediction models for forecasting. This work is largely inspired by \citet{tian2021drought} in the modeling of univariate point-valued time series. Specifically, we first introduce an auto-segmentation and clustering algorithm for ITS based on block Toeplitz sparse precision matrix estimation. Each obtained cluster is treated as a class, and each segment is imaged using Joint Recurrence Plot (JRP) \citep{eckmann1995recurrence}, resulting in an image dataset containing multiple classes. Then, we select an appropriate neural network to train on this image dataset for constructing a feature extraction network. By utilizing this feature extraction network to extract representations of the raw data and input them into traditional prediction models, we aim to achieve the forecasting of the raw data. In terms of optimizatio, we utilize the majorization-minimization algorithm to transform the highly non-convex problem of automatic segmentation and clustering into two subproblems. We provide efficient solution methods for the two subproblems based on alternating direction method of multipliers (ADMM) \citep{boyd2011distributed} and dynamic programming. Theoretically, we prove the convergence of block Toeplitz sparse precision matrix estimation. Additionally, we validate the effectiveness of the proposed method through large-scale ITS forecasting in financial scenarios.

The remainder of the paper is organized as follows. Section \ref{sec2} presents the proposed feature extraction procedure, which includes auto-segmentation and clustering of ITS, multivariate time series imaging, and the construction of the feature extraction network. Section \ref{sec3} focuses on optimization aspects, encompassing the estimation of assignment set and block Toeplitz sparse precision matrices, as well as the selection of hyperparameters. Section \ref{sec4} presents theoretical results regarding algorithm convergence. Section \ref{sec5} showcases the modeling results for stock market data and cryptocurrency data. Section \ref{sec6} summarizes the entire paper. The proofs of all theoretical results are provided in Appendix \ref{appendix}.

\section{The feature extraction procedure}\label{sec2}
In this section, we present the proposed feature extraction procedure for large-scale ITS, including the auto-segmentation and clustering algorithm, the multivariate time series imaging method, and the construction of the feature extraction network.

\subsection{Auto-segmentation and clustering of ITS} \label{sec21}

Given an ITS \((y_t)^T_{t=1}\) of length \(T\) and dimension \(n\), our first step is to automatically segment it into equal-width subsequence while performing clustering, where \(y_t = (y_{1,t}, y_{2,t}, \cdots, y_{n,t})^\top\) is the \(t\)-th observation. In \((y_t)^T_{t=1}\), \(y_{i,j} = \left[y^l_{i,j}, y^u_{i,j}\right], 1\leq i \leq n, 1\leq t \leq T\) represents an interval , where \(y^l_{i,j}\) and \(y^u_{i,j}\) are the upper and lower bounds of the interval \(y_{i,j}\), respectively, satisfying \(y^l_{i,j} \leq y^u_{i,j}\). In our setting, the dimension \(n\) may be relatively large.

Clustering methods for point-valued time series can be broadly categorized into model-based and distance-based approaches \citep{DBLP:conf/ijcai/HallacVBL18}. However, these methods have not been extended to interval-valued scenarios and are typically designed for individual observations rather than subsequences. Therefore, we need to develop clustering methods specifically for ITS subsequences. There are three reasons for performing clustering on subsequences: (\romannumeral1) the subsequent imaging method JRP is designed for subsequences; (\romannumeral2) our feature extraction and prediction models are built on subsequences; and (\romannumeral3) subsequences contain more pattern information. The first two reasons are straightforward to understand, we now provide more insights into the last one through example. In the process of car driving, its states, such as acceleration, deceleration, constant speed, and standstill, typically occur within a certain time window and are unlikely to change rapidly. Therefore, representing a state with observations over a period of time is more appropriate.

Before clustering subsequences, we first construct a one-to-one mapping of ITS \((y_t)^T_{t=1}\). Specifically, given a window width \(w\), we let observations \(y_{t-w+1}, y_{t-w+2}, \cdots, y_{t}\) ending at \(t\) constitute a subsequence. In this way, we can obtain a total of \(T\) subsequences. By vectorizing each subsequence, we obtain an interval vector of dimension \(nw\), where the \(t\)-th interval vector is \(Y_t = \text{vec}(y_{t-w + 1}, y_{t-w+2}, \cdots, y_t), \ t = 1,2,\cdots, T\), and \(\text{vec}(\cdot)\) is the vectorization operator. It is worth noting that since the ITS starts from \(y_1\), the first \(w\) interval vectors are shorter. At this point, \(y_t\) and \(Y_T,\ t = 1,2,\cdots, T\) are one-to-one mappings. The segmentation window width \(w\) is a hyperparameter, typically set to \(w \ll T\). We discuss the impact of its value on the forecasting performance in Section \ref{sec5}.

In the following, our goal is to cluster the \(T\) interval vectors \((Y_t)^T_{t=1}\) of dimension \(nw\) into \(K\) clusters. Our method is model-based, where each cluster has different structural features, such as (conditional) dependency structures. Following \citet{DBLP:conf/ijcai/HallacVBL18}, we use precision matrices (inverse covariance matrices) to describe the structural information of each cluster. It is worth noting that our clustering objects are different from those of \citet{DBLP:conf/ijcai/HallacVBL18}, hence the subsequent optimization problem and its solution are completely different as well. The covariance matrix can capture the marginal correlations between variables. The precision matrix can capture the conditional correlations, i.e., the correlations between pairs of variables given the remaining variables, which is closely related to undirected graphs under a Gaussian model \citep{Fanoverview2016}. 

Let the precision matrix \(\mTheta_k = (\theta^k_{i,j})_{nw \times nw}\) describe the structural information of the \(k\)-th cluster. Then, our clustering process is equivalent to estimating
\[
(\mTheta_k)^K_{k=1},\ (P_k)^K_{k=1},	
\]
where \((P_k)^K_{k=1}\) is the assignment set satisfying \(P_k \subset \{1,2,\cdots, T\}\) and \(P_k \cap P_j = \varnothing\) for \(k \neq j\). Since the dimensionality \(n\) of ITS is typically high, to ensure that the estimated precision matrices has excellent asymptotic properties, it is often necessary to impose certain structural assumptions during the estimation process, such as sparsity. Classical penalized likelihood methods in high-dimensional statistics can efficiently and conveniently estimate precision matrices \citep{Fanoverview2016}. However, there is currently no unified definition for the covariance matrix of interval-valued data, let alone the precision matrix \citep{tian2024minimum}. Therefore, in this paper, we assume from an intuitive perspective that the upper and lower bounds of intervals exhibit consistent dependency structures. For instance, generally, when the daily high price of a stock increases, its daily low price also tends to increase. Without loss of generality, we additionally assume that the upper and lower bounds of interval vectors within each cluster are independently and identically distributed (i.i.d.) according to a normal distribution. Altogether, these assumptions can be formalized as
\[
(Y^l_t)_{\in P_k} \overset{\text{i.i.d}}{\sim} \mathcal{N}(\vmu^l_k, \mSigma_k), \ (Y^u_t)_{\in P_k} \overset{\text{i.i.d}}{\sim} \mathcal{N}(\vmu^u_k, \mSigma_k),\ k = 1,2,\cdots,K,
\]
where \(Y^l_t\) and \(Y^u_t\) are respectively the upper and lower bounds of interval vectors \(Y_t\), \(\vmu^l_k\) and \(\vmu^u_k\) are respectively the mean of the upper and lower bounds, and \(\mSigma_k = \mTheta^{-1}_k, k = 1,2,\cdots, K\) are the covariance matrices.

Directly, the negative log-likelihood function corresponding to the samples in the \(k\)-th cluster can be written in terms of the upper and lower bounds, which are defined as follows,
\begin{equation} \label{negativeloglikelihood}
\begin{aligned} 
\ell^l(Y^l_t, \mTheta_k) &= (1/2) (Y^l_t -\vmu^l_k)^\top \mTheta_k(Y^l_t -\vmu^l_k) -(1/2) \log \det\mTheta_k +(n/2) \log(2\pi),\\
\ell^u(Y^u_t, \mTheta_k) &=(1/2) (Y^u_t - \vmu^u_k)^\top \mTheta_k(Y^u_t - \vmu^u_k) -(1/2) \log \det\mTheta_k + (n/2) \log(2\pi).\\
\end{aligned}
\end{equation}

When the data do not follow a normal distribution or exhibit dependence, we refer to (\ref{negativeloglikelihood}) as a pseudo log-likelihood function. Similar to the classical methods for estimating the precision matrix, we propose a modified penalized likelihood method to estimate the precision matrix of the \(k\)-th cluster, defined as
\begin{equation} \label{penalizedlike}
\widehat{\mTheta}_k = \argmin_{\mTheta_k \in \gT} \sum_{i \neq j}p_{\lambda_k}(\vert \theta^k_{i,j} \vert)  + \sum_{t\in P_k} \ell^l(Y^l_t, \mTheta_k) + \ell^u(Y^u_t, \mTheta_k),
\end{equation}
where \(p_{\lambda_k}(\cdot)\) is a penalty function with turning parameter \(\lambda_k\) that promotes the sparsity of the precision matrix, \(\gT\) is a set of symmetric block Toeplitz matrices with dimension \(nw \times nw\). Commonly used penalty functions include Lasso \citep{tibshirani1996regression}, adaptive Lasso \citep{zou2006adaptive}, smoothly clipped absolute deviation (SACD) \citep{JianqingVariable2001}, and minimax concave penalty (MCP) \citep{10.1214/09-AOS729}. It is worth noting that the optimization objective (\ref{penalizedlike}) includes both upper and lower bound likelihoods, which is different from point-valued data.

Since we need to estimate the precision matrices \((\mTheta_k)^K_{k=1}\) and assignment set \((P_k)^K_{k= 1}\) corresponding to \(K\) clusters simultaneously, our overall optimization objective is
\begin{equation} \label{optall}
\argmin_{(\mTheta_k)^K_{k=1} \subset \mathcal{T}, (P_k)^K_{k=1}} \sum_{k=1}^{K} \left(
\sum_{i \neq j} p_{\lambda_k}(\vert \theta^k_{i,j} \vert) +  \sum_{t\in P_k} \ell^l(Y^l_t, \mTheta_k) + \ell^u(Y^u_t, \mTheta_k) + \beta \bm{1}(Y_{t-1} \notin P_k)
\right),
\end{equation}
where \(\beta\) is a penalty coefficient that encourages temporal consistency, and a larger \(\beta\) results in neighboring samples belonging to the same cluster, \(\bm{1}(\cdot)\) is the indicator function. Temporal consistency refers to encouraging adjacent samples to be assigned to the same cluster. The reason for adding the temporal consistency constraint to the overall optimization objective (\ref{optall}) is that the state of the system generally does not undergo abrupt changes (as in the previously mentioned case of car driving).

In the following, we explain why we assume that the precision matrix is block Toeplitz. A block Toeplitz matrix has the following structure, 
\[
\mTheta = 
\left[
\begin{array}{cccccc}
C^{(0)} &  (C^{(1)})^\top &  (C^{(2)})^\top & \cdots  &\cdots  & (C^{(w-1)})^\top \\
C^{(1)} &  C^{(0)} &  (C^{(1)})^\top & \ddots  &   & \vdots \\
C^{(2)} &  C^{(1)} &  \ddots & \ddots  &\ddots  & \vdots \\ 
\vdots & \ddots &  \ddots & \ddots &  (C^{(1)})^\top & (C^{(2)})^\top \\
\vdots & & \ddots & C^{(1)} &  C^{(0)} & (C^{(1)})^\top\\
C^{(w-1)} &  \cdots &  \cdots & C^{(2)}  &C^{(1)}  &C^{(0)} \\
\end{array}
\right],
\] 
where \(C^{(0)}, C^{(1)}, \cdots, C^{(w-1)}\in \mathbb{R}^{n\times n}\) are sub-block matrices, and \(\mTheta\) is determined by these matrices. The sub-block matrices on the diagonal describe the conditional dependencies within time. For example, \(C^{(0)}_{i,j}\) represents the dependence between variables \(i\) and \(j\) at the same time. On the other hand, off-diagonal sub-block matrices describe the conditional dependencies across time. For example, \(C^{(1)}_{i,j}\) represents the dependence between variables \(i\) and \(j\) when the time difference is 1 (for example, \(t\) and \(t+1\)). Thus, the structural assumption of the precision matrix implies that the conditional dependencies between variables are time-invariant. For instance, the conditional dependencies between variables at time \(t\) and \(t+1\), as well as between \(t+1\) and \(t+2\), can both be described using the sub-block matrix \(C^{(1)}\).  

In summary, the auto-segmentation and clustering of ITS is equivalent to solving the optimization problem (\ref{optall}), that is, estimating \((\mTheta_k)^K_{k=1}\) and \((P_k)^K_{k=1}\), which is discussed in detail in Section \ref{sec3}.

\subsection{Multivariate time series imaging} \label{sec22}
Once the overall optimization problem (\ref{optall}) is solved, the precision matrices \((\mTheta_k)_{k=1}^K\) describing the structure of each cluster, along with the assignment set  \((P_k)_{k=1}^K\), become accessible. Then, each cluster is treated as a single class, and the samples within each class are transformed into images using the multivariate time-series imaging method JRP. The resulting image dataset is used for training and constructing the subsequent feature extraction network.

Below, we illustrate imaging method JRP using the subsequence \(Y_t, t\in P_k\) of the \(k\)-th cluster as an example. As JRP is designed for point-value time series, one subsequence can yield two images. Specifically, the interval vector (or subsequence) \(Y_t\) can be determined by bivariate point-value sequences \(Y^l_t, Y^u_t \in \mathbb{R}^{nw}\). Through the inverse operation of the vec operator, \(Y^l_t\) and \(Y^u_t\) can be transformed into two multivariate point-valued time series of size \(w\) and dimension \(n\), denoted as 
\[
\mY^l_t = \left[
\begin{array}{ccc}
y^l_{1, t-w+1} & \cdots & y^l_{1, t}\\
\vdots & \ddots & \vdots\\
y^l_{n, t-w+1} & \cdots & y^l_{n, t}\\
\end{array}
\right], \ \mY^u_t = \left[
\begin{array}{ccc}
y^u_{1, t-w+1} & \cdots & y^u_{1, t}\\
\vdots & \ddots & \vdots\\
y^u_{n, t-w+1} & \cdots & y^u_{n, t}\\
\end{array}
\right], 
\]
respectively. Then, we utilize JRP to transform \(\mY^l_t\) and \(\mY^u_t\) into images. If we directly cluster the center and range subsequences of the intervals, imaging can also be conducted in this manner. Of course, there are also studies that directly image multivariate interval-valued time series. For example, \citet{tianqindk2024} extended JRP to interval scenarios based on suitable distance measures between paired intervals. In this paper, we only consider simple point-value time series imaging method. This approach yields double the number of images, which is beneficial for constructing and training feature extraction networks.

JRP is a multivariate extension of the univariate point-valued time series imaging method, Recurrence Plot (RP) \citep{eckmann1995recurrence}. JRP first utilizes RP to transform each dimension of the multivariate time series into an image, which is then fused. For example, for the \(n\)-th dimension observation \((y^l_{n, t-w+1},y^l_{n, t-w+2},\cdots, y^l_{n, t})^\top\) of the multivariate point-valued time series \(\mY^l_t\) composed of upper bounds, RP first defines the trajectory of length \(\kappa\) as
\[
\vec{y}^l_{n, i} = \left(y^l_{n, i}, y^l_{n, i+\kappa},\cdots, y^l_{n,i+(m-1)\kappa}\right)^\top, \ i = t-w+1,\cdots, t -(m-1)\kappa, 
\]
where \(\kappa\) represents the time gap between two adjacent points in the trajectory, and \(m\) denotes the dimension of the trajectory. Let \(R^l_n\) be the image obtained using the RP method, then we have
\[
(R^l_{n})_{i,j} = H(\epsilon_n - \lVert \vec{y}^c_{n, i} -  \vec{y}^c_{n, j}\rVert),\ i,j = t-w+1,\cdots, t -(m-1)\kappa,
\]
where \(H(\cdot)\) is the Heaviside function, \(\epsilon_n\) is the threshold for the \(n\)-th dimension, and \(\lVert \cdot \rVert\) is the Euclidean norm. Similarly, we can obtain images of other dimensions, namely \(R^l_{1}, R^l_{2}, \cdots, R^l_{n-1}\). JRP utilizes the Hadamard product \(\odot\) to fuse the images of each dimension, resulting in the image \(J^l_t\) corresponding to \(\mY^l_t\), with
\[
J^l_t = R^l_1 \odot R^l_2 \odot \cdots \odot R^l_n.
\]

Similarly, we can obtain the image \(J^u_t\) corresponding to \(\mY^u_t\). The definition of JRP indicates that each RP shares the same dimension, i.e., all trajectories are of the same size. Additionally, as the threshold values \(\epsilon_1, \epsilon_2, \cdots, \epsilon_n\) of each dimension vary, the pattern information exhibited by JRP also varies. For example, if \(\max_h\epsilon_h < \min_{h, i, j}\lVert \vec{y}^l_{h, i} - \vec{y}^l_{h, j}\rVert\) or \(\min_h\epsilon_h \geq \max_{h, i, j}\lVert \vec{y}^l_{h, i} - \vec{y}^l_{h, j}\rVert\), JRP become an all-zero matrix or an all-one matrix, respectively. In these cases, the JRP scarcely demonstrates any information about the multivariate time series. Selecting these threshold values is relatively crucial, and they are generally chosen as the quantile of the distances. Algorithm \ref{imagingdatap} outlines the procedure for constructing an image dataset comprising \(K\) classes.
\begin{algorithm}[H]
\caption{Procedure for constructing a dataset comprising \(K\) classes}
\label{imagingdatap}
\begin{algorithmic}
\REQUIRE ~~\\
Assignment set \((P_k)^K_{k=1}\),\\ 
Threshold values \((\epsilon_h)^n_{h=1}\),\\
Dimensions of the trajectory \((m_k)^K_{k=1}\),\\
Subsequences \((Y_t)^T_{t=1}\).
\ENSURE ~~\\
Representing \((Y_t)^T_{t=1}\) as \((Y^l_t)^T_{t=1}\) and \((Y^u_t)^T_{t=1}\).\\
\textbf{for} \(k =1,2,\cdots, K\)\\
\quad \quad \textbf{for} \(t = 1,2,\cdots, \vert P_k \vert\)\\
\quad \quad \quad \quad Using JRP transform \(\mY^l_t\) and \(\mY^u_t\) into images \(J^l_t\) and \(J^u_t\), respectively.\\
\quad \quad \quad \quad Assign class labels \(k\) to the images.\\
\quad \quad \textbf{end}\\
\ \textbf{end}\\
\textbf{Output}: Image dataset \(\mathcal{D} = \left(\left((J^l_t, k),(J^u_t, k)\right)_{t\in P_k}\right)^K_{k=1}\) comprising \(K\) classes.
\end{algorithmic}
\end{algorithm}

\subsection{The construction of feature extraction network}\label{sec23}
Through the first two steps (Sections \ref{sec21} and \ref{sec22}), we obtain an image dataset \(\mathcal{D}\) comprising \(K\) classes. With this dataset, we train a feature extraction network for subsequent feature extraction on large-scale ITS. In constructing the feature extraction network, we employ the concept of transfer learning. Specifically, the source task is image classification, while the target task is forecasting. Since the source task and the target task use the same ITS dataset, if a neural network achieves sufficiently high classification accuracy on the image dataset \(\mathcal{D}\), we can infer that the neural network has an excellent deep representation of the raw data. Consequently, the features extracted based on this neural network can be effectively used for forecasting. We refer to such a neural network as a feature extraction network.

\begin{figure}[H]
\centering 
\includegraphics[scale=0.56]{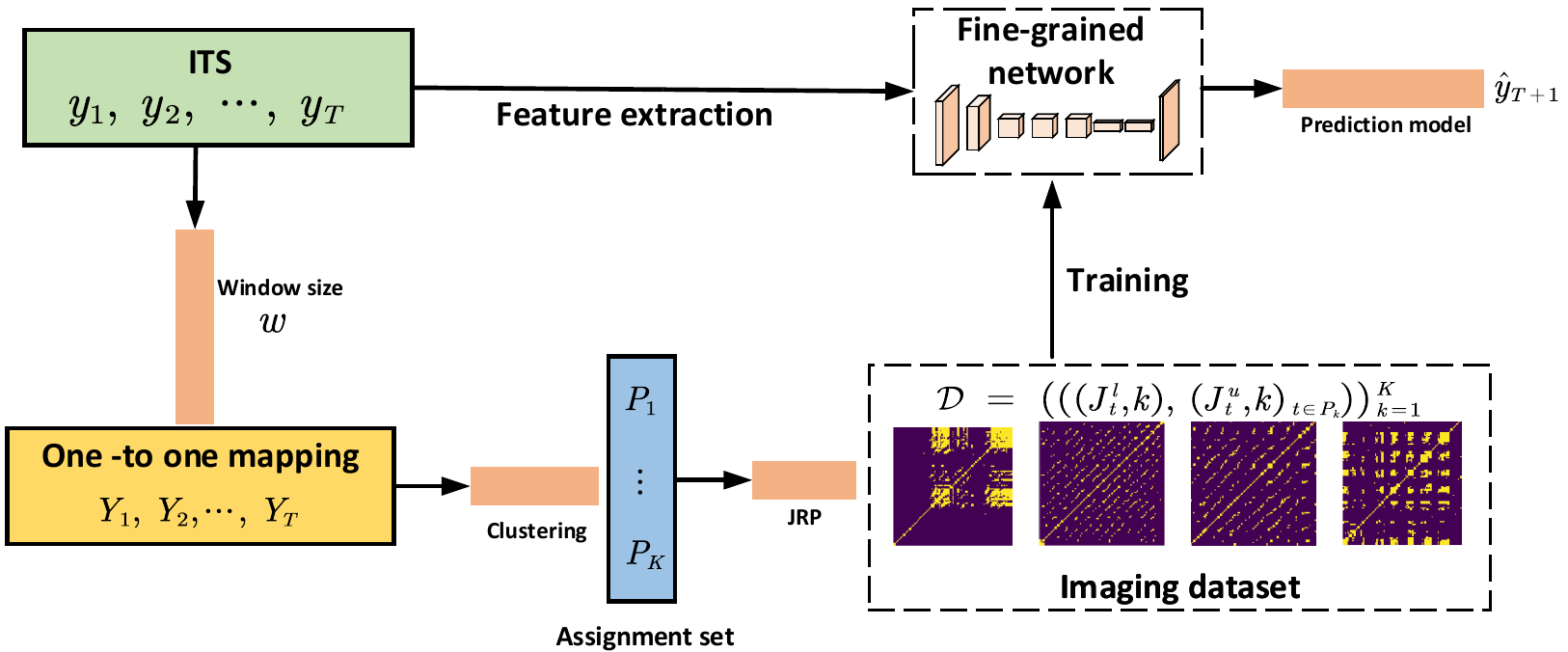}
\caption{The technical roadmap of the proposed method includes auto-segmentation and clustering, multivariate time series imaging, feature extraction network training, feature extraction, and forecasting.}\label{technology_roadmap}
\end{figure}

The choice of neural network architecture significantly impacts the subsequent forecasting performance. Some popular network architectures, such as ResNet \citep{DBLP:conf/cvpr/HeZRS16} and VGG \citep{DBLP:journals/corr/SimonyanZ14a} with varying layers, may achieve high classification accuracy; however, the deep representations obtained from these models may not perform well for forecasting. We discuss this aspect in detail in Section \ref{sec5}. Finally, Figure \ref{technology_roadmap} presents the technical roadmap of the proposed method.

\section{The optimization aspect}\label{sec3}
As discussed in Section \ref{sec3}, estimating the precision matrices \((\mTheta_k)^K_{k=1}\) and the assignment set \((P_k)^K_{k=1}\), which involves solving optimization problem (\ref{optall}), is a crucial step in constructing the feature extraction network. Since optimization problem (\ref{optall}) involves a mixture of continuous and combinatorial optimization, comprising two sets of parameters, \((\mTheta_k)^K_{k=1}\) and \((P_k)^K_{k=1}\), which is highly non-convex. As there are no tractable methods to obtain the global optimal solution, we adopt the majorization-minimization algorithm to update the assignment set and precision matrices alternately.

This entails fixing one set while solving for the other until convergence is achieved. Additionally, the selection of window width \(w\), the number of clusters \(K\), and the regularization parameters \((\lambda_k)^K_{k=1}\) in the penalty function are also discussed.

\subsection{Estimation of assignment set}
In this section, we fix the precision matrices \((\mTheta_k)^K_{k=1}\) and estimate the assignment set \((P_k)^K_{k=1}\). Removing some constant terms, this involves solving the following optimization problem:
\begin{equation}\label{Estep}
\argmin_{(P_k)^K_{k=1}} \sum_{k=1}^{K} \left(\sum_{t\in P_k} \ell^l(Y^l_t, \mTheta_k) + \ell^u(Y^u_t, \mTheta_k) + \beta \bm{1}(Y_{t-1} \notin P_k)\right).
\end{equation}

Optimization problem (\ref{Estep}) aims to minimize the negative log-likelihood function and enforce temporal consistency simultaneously while assigning \(T\) subsequences to \(K\) clusters. The parameter \(\beta\) serves as the regularization parameter to balance the two objectives. If \(\beta = 0\), this implies that there is no penalty for temporal consistency, and all subsequences will be assigned to \(K\) clusters independently. In this case, solving the problem (\ref{Estep}) simply involves minimizing the negative log-likelihood function. As \(\beta \to \infty\), the penalty for temporal consistency becomes so large that all subsequences will be assigned to exactly one cluster. The proper choice of the regularization parameter \(\beta\) is problem-specific. We discuss the impact of different values of \(\beta\) on forecasting performance in Section \ref{sec5}.

Although problem (\ref{Estep}) is combinatorial, involving the assignment of \(T\) subsequences to \(K\) clusters, the Viterbi algorithm \citep{viterbi1967error} can efficiently find the global optimal solution, with a computational complexity of only \(\mathcal{O}(KT)\). Before discussing the solution to optimization problem (\ref{Estep}), we first introduce some necessary notations. Let \(i_1,i_2,\cdots,i_T\) denote the clusters in which the subsequences \((Y_t)^T_{t=1}\) are located, respectively. The objective function in (\ref{Estep}) is for all \(T\) subsequences. Then, we have the partial objective function for the first \(t\) sequences \(Y_1,Y_2,\cdots, Y_t\) as
\[
\mathcal{A}(i_1,i_2,\cdots,i_t) = \sum_{h= 1}^{t}\ell^l(Y^l_h, \mTheta_{i_h}) + \ell^u(Y^u_t, \mTheta_{i_h}) + \beta \bm{1}(Y_{l-1} \notin P_{i_h}),\ t = 1,2,\cdots, T,
\]
and it is evident that \(\mathcal{A}(i_1,i_2,\cdots,i_T)\) is equivalent to the objective function in (\ref{Estep}). 

Define the minimum partial objective function \(\mathcal{A}(i_1,i_2,\cdots,i_t)\) among all paths \((i_1,i_2,\cdots, i_t)\) with the subsequence \(Y_t\) assigned to the \(k\)-th cluster as
\[
\delta_t(k) = \min_{i_1,i_2,\cdots,,i_{t-1}}\mathcal{A}(i_1,i_2,\cdots,i_{t-1}, i_t = k), \ k= 1,2,\cdots,K.
\]

From the definition of \(\delta_t(k)\),  we have
\begin{equation} \label{delta}
\begin{aligned} 
\delta_{t+1}(k) &= \min_{i_1,i_2,\cdots,i_t}\mathcal{A}(i_1,i_2,\cdots,i_t,i_{t+1} = k)\\
&=\min_{i_1,i_2,\cdots,i_{t-1}} \min_{1\leq j \leq K}\mathcal{A}(i_1,i_2,\cdots,i_{t-1}, i_t = j, i_{t+1} = k)\\
&= \min_{1\leq j \leq K}\min_{i_1,i_2,\cdots,i_{t-1}}\mathcal{A}(i_1,i_2,\cdots,i_{t-1}, i_t = j) + \ell^l(Y^l_{t+1}, \mTheta_{k}) + \ell^u(Y^u_{t+1}, \mTheta_{k}) + \beta \bm{1}(Y_{t} \notin P_{k})\\
&= \min_{1\leq j \leq K} \delta_t(j) + \ell^l(Y^l_{t+1}, \mTheta_{k}) + \ell^u(Y^u_{t+1}, \mTheta_{k}) + \beta \bm{1}(Y_{t} \notin P_{k}),
\end{aligned} 
\end{equation}
where \(k = 1,2,\cdots, K, t = 1,2,\cdots, T-1\). When \(Y_t\) is in the \(k\)-th cluster, let \(\Psi_t(k)\) be the cluster to which \(Y_{t-1}\) is assigned among the paths \((i_1,i_2,\cdots,i_{t-1}, k)\) with the minimum partial objective function \(\mathcal{A}(i_1,i_2,\cdots,i_{t-1}, k)\). We have
\begin{equation} \label{psi}
\Psi_t(k) = \argmin_{1\leq j \leq K} \delta_{t-1}(j) + \ell^l(Y^l_{t}, \mTheta_{k}) + \ell^u(Y^u_{t}, \mTheta_{k}) + \beta \bm{1}(Y_{t-1} \notin P_{k}), \ k = 1,2,\cdots, K.
\end{equation}

Using the definitions of \(\delta\) and \(\Psi\), we present the solution procedure for optimization problem (\ref{Estep}) in Algorithm \ref{Viterbipath}.

\begin{algorithm}[H]
\caption{Solving procedure of problem (\ref{Estep}) using Viterbi algorithm}
\label{Viterbipath}
\begin{algorithmic}
\REQUIRE ~~\\
Precision matrices \((\mTheta_{k})^K_{k=1}\),\\ 
Subsequences \((Y_t)^T_{t=1}\).
\ENSURE ~~\\
\textbf{(1) Initialization} 
\[
\begin{aligned}
	\delta_1(k) &= \mathcal{A}(i_1 = k), \ k = 1,2,\cdots,K,\\
	\Psi_1(k) &= 0,\  k = 1,2,\cdots, K.
\end{aligned}
\]
\textbf{(2) Traversal} \\
\qquad \textbf{for} \(t =2,3,\cdots, T\)\\
\[
\begin{aligned}
\delta_{t}(k) &= \min_{1\leq j \leq K} \delta_{t-1}(j) + \ell^l(Y^l_{t}, \mTheta_{k}) + \ell^u(Y^u_{t}, \mTheta_{k}) + \beta \bm{1}(Y_{t-1} \notin P_{k}), \ k=1,2,\cdots,K \\
\Psi_t(k) &= \argmin_{1\leq j \leq K} \delta_{t-1}(j) + \ell^l(Y^l_{t}, \mTheta_{k}) + \ell^u(Y^u_{t}, \mTheta_{k}) + \beta \bm{1}(Y_{t-1} \notin P_{k}),\ k=1,2,\cdots,K, 	
\end{aligned}
\]
\ \qquad \textbf{end}\\
\textbf{(3) Termination} 
\[
\begin{aligned}
	A^* &= \min_{1\leq k \leq K} \delta_{T}(k),\\
	i^*_T &= \argmin_{1\leq k \leq K} \delta_{T}(k).
\end{aligned}
\]
\textbf{(4) Optimal Path Retrieval}
\[
i^*_t = \Psi_{t+1}(i^*_{t+1}), \ t = T-1,T-2,\cdots, 1.
\]
\textbf{Output}: optimal path \((i^*_1,i^*_2,\cdots, i^*_T)\).
\end{algorithmic}
\end{algorithm}

Based on the optimal path \((i^*_1,i^*_2,\cdots, i^*_T)\) output by Algorithm \ref{Viterbipath} (indicating the assignment of each subsequence to a cluster), we can immediately obtain the optimal assignment set when the precision matrices are fixed.

\subsection{Estimation of the precision matrices}
In the following, we discuss how to estimate the precision matrices \((\mTheta_k)^K_{k=1}\) with a fixed assignment set \((P_k)^K_{k=1}\). In this case, optimization problem (\ref{optall}) is equivalent to
\begin{equation}\label{optallEstep}
\argmin_{(\mTheta_k)^K_{k=1} \subset \mathcal{T}} \sum_{k=1}^{K} \left(
\sum_{i \neq j} p_{\lambda_k}(\vert \theta^k_{i,j} \vert) +  \sum_{t\in P_k} \ell^l(Y^l_t, \mTheta_k) + \ell^u(Y^u_t, \mTheta_k)
\right).
\end{equation}

Through simple algebraic operations, we have
\[
\sum_{t \in P_k} \ell^l(Y^l_t, \mTheta_k) + \ell^u(Y^u_t, \mTheta_k) = |P_k|(\text{Tr}(\mS^l_k \mTheta_{k}) + \text{Tr}(\mS^u_k \mTheta_{k}) - 2\log \det \mTheta_{k})+C, \\
\]
where \(|P_k|\) represents the number of subsequences in cluster \(P_k\), \(C\) is a constant independent of \(\mTheta_{k}\), \(\mS^l_k\) and \(\mS^u_k\) are the empirical covariance matrices corresponding to the upper and lower bound subsequences. From the optimization objective (\ref{optallEstep}), it can be observed that we can estimate the \(K\) precision matrices in parallel. Ignoring the subscript \(k\) indicating clusters, we need to simultaneously solve \(K\) optimization problems of the following form:

\begin{equation} \label{Mstep}
\begin{aligned}
&\min \text{Tr}(\mS^l \mTheta) + \text{Tr}(\mS^u \mTheta)- 2\log \det \mTheta +  (1/|P|) \sum_{i \neq j} p_\lambda(\vert \theta_{i,j} \vert),\\
& \text{subject to } \mTheta \in \mathcal{T}.
\end{aligned}
\end{equation}

To utilize the ADMM algorithm for solving optimization problem (\ref{Mstep}), we introduce a consensus variable \(\mGamma\) and rewrite problem (\ref{Mstep}) into its equivalent form as follows:

\begin{equation} \label{Mstepfisrststage}
\begin{aligned}
&\min \text{Tr}(\mS^l \mTheta) + \text{Tr}(\mS^u \mTheta)- 2\log \det \mTheta + (1/|P|)\sum_{i \neq j} p_\lambda(\vert \theta_{i,j} \vert),\\
& \text{subject to } \mTheta = \mGamma, \mGamma \in \mathcal{T}.
\end{aligned}
\end{equation}

The augmented Lagrangian function corresponding to optimization problem (\ref{Mstepfisrststage}) is
\[
\begin{aligned}
L(\mGamma, \mTheta;\mLambda) &= \text{Tr}(\mS^l \mTheta) + \text{Tr}(\mS^u \mTheta)- 2\log \det \mTheta + (1/|P|)\sum_{i \neq j} p_\lambda(\vert \theta_{i,j} \vert) \\&+ \langle \mLambda, \mTheta-\mGamma \rangle + (1/2\rho)\lVert \mTheta - \mGamma \rVert^2_F,
\end{aligned}
\]
where \(\mLambda\) represents the Lagrangian multiplier matrix, and \(\rho\) is a given penalty parameter. Assuming that the estimates obtained in the \(q\)-th iteration are \(\mGamma^{(q)}\), \(\mTheta^{(q)}\), and \(\mLambda^{(q)}\), the ADMM algorithm in the (\(q+1\))-th iteration consists of the following three steps:
\begin{align}
&\mGamma \text{ step:  } \mGamma^{(q+1)} = \argmin_{\mGamma \in \mathcal{T}} L(\mTheta^{(q)}, \mGamma;\mLambda^{(q)}), \label{mGammastep}\\
&\mTheta \text{ step:  }  \mTheta^{(q+1)} = \argmin_{\mTheta} L(\mTheta, \mGamma^{(q+1)};\mLambda^{(q)}), \label{mThetastep}\\
&\mLambda \text{ step:  } \mLambda^{(q+1)} = \mLambda^{(q)} - (1/\rho)(\mTheta^{(q+1)} -\mGamma^{(q+1)}). 
\end{align}

In each iteration, we execute the above three steps until convergence. In the following, we specifically discuss the solution to the optimization problems in \(\mGamma\) step (\ref{mGammastep}) and \(\mTheta\) step (\ref{mThetastep}).  

For \(\mGamma\) step (\ref{mGammastep}), we have

\begin{equation}\label{Tstep1}
\mGamma^{(q+1)} = \argmin_{\mGamma \in \mathcal{T}} L(\mTheta^{(q)}, \mGamma;\mLambda^{(q)})  = \argmin_{\mGamma \in \mathcal{T}}\langle \mLambda^{(q)}, \mTheta^{(q)}-\mGamma \rangle + (1/2\rho)\lVert \mTheta^{(q)} - \mGamma \rVert^2_F. 
\end{equation}

Since \(\mGamma\) is a symmetric block Toeplitz matrix, \(C^{(0)}\) is symmetric. In optimization problem (\ref{Tstep1}), we can solve each sub-block matrices \(C^{(0)}\), \(C^{(1)},\cdots, C^{(w-1)}\) in parallel. Furthermore, within each sub-block matrix, we can solve for each element individually. For matrix \(C^{(0)}\), we need to estimate \(n(n+1)/2\) elements individually, and for the off-diagonal sub-block matrices, we need to estimate \(n^2\) elements. Thus, optimization problem (\ref{Tstep1}) is equivalent to \(n(n+1)/2 + (w-1)n^2\) independent sub-problems, each with the same form and solvable in the same manner. Let \(D\) denote the number of occurrences of an element in matrix \(\mGamma\). For the sub-block matrices \(C^{(d)}, d = 0,1,\cdots, w-1\) (where \(d=0\) corresponds to the off-diagonal elements of \(C^{(0)}\)), we have \(D=2(w-d)\); for the diagonal elements of \(C^{(0)}\) (which are also the diagonal elements of \(\mGamma\)), we have \(D=w\). 

Let \(\mathcal{B}^{(d)}_{i,j} = (\mathcal{B}^{(d)}_{i,j, g})^D_{g=1}\), where \(\mathcal{B}^{(d)}_{i,j, g}\) represents the \(g\)-th occurrence of the \((i,j)\)-th element in \(C^{(d)}\) as indexed in \(\mGamma\). Consequently, the elements in matrix \(\mGamma\) indexed in \(\mathcal{B}^{(d)}_{i,j}\) share the same value, all equal to
\[
\argmin_z \sum_{g=1}^{D} -\mLambda^{(q)}_{i,j, g} z + (1/2\rho) (\mTheta^{(q)}_{i,j, g} -z)^2,
\]
which has the following closed-form solution
\begin{equation}\label{Tclosr}
z = \left(\sum_{g=1}^{D} \mTheta^{(q)}_{i,j, g} + \rho \mLambda^{(q)}_{i,j, g}\right) \bigg/ D.
\end{equation}

By solving \(n(n+1)/2 + (w-1)n^2\) instances of the above problem in parallel, we can obtain an estimate for \(\mGamma\). Regarding the choice of the penalty function, since Lasso \citep{tibshirani1996regression} is biased and may result in inconsistencies, we use SCAD, proposed by \citet{JianqingVariable2001}, for estimation. Its form is:
\[
p_\lambda(|x|) = \int_{0}^{|x|} \lambda\bm{1}(b\leq \lambda) + \frac{(a\lambda - b)_+}{a - 1} \bm{1}(b > \lambda)db, 
\]
where \(\lambda\) is a tuning parameter, and \(a > 2\) is a clipped constant, and
\citet{JianqingVariable2001} suggest setting \(a = 3.7\) from a Bayesian risk minimization perspective. Using local linear approximation (LLA) \citep{10.1214/009053607000000802}, the optimization objective of the \(\mTheta\) step (\ref{mThetastep}) in the (\(q+1\))-th iteration can be approximated as
\[
\begin{aligned}
L(\mTheta, \mGamma^{(q+1)};\mLambda^{(q)}) &= \text{Tr}(\mS^c \mTheta) + \text{Tr}(\mS^r \mTheta)- 2\log \det \mTheta + (1/|P|)\sum_{i \neq j} p_\lambda(\vert \theta_{i,j} \vert) \\
&+ \langle \mLambda^{(q)}, \mTheta-\mGamma^{(q+1)} \rangle + (1/2\rho)\lVert \mTheta - \mGamma^{(q+1)} \rVert^2_F\\
& \approx \text{Tr}(\mS^c \mTheta) + \text{Tr}(\mS^r \mTheta)- 2\log \det \mTheta + (1/|P|)\sum_{i \neq j} p_\lambda(\vert \theta^{(q)}_{i,j} \vert) + \dot{p}_\lambda(\vert \theta^{(q)}_{i,j} \vert) (|\theta_{i,j}| - \theta^{(q)}_{i,j}) \\
&+ \langle \mLambda^{(q)}, \mTheta-\mGamma^{(q+1)} \rangle + (1/2\rho)\lVert \mTheta - \mGamma^{(q+1)} \rVert^2_F\\
& = \text{Tr}(\mS^c \mTheta) + \text{Tr}(\mS^r \mTheta)- 2\log \det \mTheta + (1/|P|)\sum_{i \neq j} \dot{p}_\lambda(\vert \theta^{(q)}_{i,j} \vert) |\theta_{i,j}| \\
&+ \langle \mLambda^{(q)}, \mTheta-\mGamma^{(q+1)} \rangle + (1/2\rho)\lVert \mTheta - \mGamma^{(q+1)} \rVert^2_F,\\
\end{aligned}
\]
where \(\dot{p}_\lambda(\cdot)\) is the derivative of the SCAD penalty, and \(\mTheta^{(q)} =(\theta^{(q)}_{i,j})_{nw \times nw}\) is the estimate obtained in the \(q\)-th iteration. Defining the matrix \(\mG^{(q)}\) such that \((\mG^{(q)})_{i,j} = (1/|P|) \dot{p}_\lambda(|\theta^{(q)}_{ij}|)\bm{1}(i \neq j)\). Then, we have \(L(\mTheta, \mGamma^{(q+1)};\mLambda^{(q)})\) approximately equivalent to
\begin{equation}\label{LLAappro}
\text{Tr}(\mS^c \mTheta) + \text{Tr}(\mS^r \mTheta)- 2\log \det \mTheta + \mG^{(q)} \odot \lVert \mTheta\rVert_1  + \langle \mLambda^{(q)}, \mTheta-\mGamma^{(q+1)} \rangle + (1/2\rho)\lVert \mTheta - \mGamma^{(q+1)} \rVert^2_F, 
\end{equation}	
which is equivalent to a graphical elastic net \citep{kovacs2021graphical}. According to the first-order optimality condition, setting the partial derivative of (\ref{LLAappro}) with respect to \(\mTheta\) to zero, we have
\begin{equation} \label{partialdev}
\mS^c + \mS^r - 2\mTheta^{-1} + \mG^{(q)} \odot \mA + \mLambda^{(q)} + (1/\rho)(\mTheta - \mGamma^{(q+1)}) =0,
\end{equation}
where \(\mA = (a_{i,j})_{nw \times nw}\) takes the following form
\[
a_{i,j} \begin{cases}
= \sign(\theta_{i, j})\quad \theta_{i,j} \neq 0\\
\in [-1,1]\quad \text{otherwise},
\end{cases}
\]
with \(\sign(\cdot)\) as the sign function. Let \(\mW\coloneqq \mTheta^{-1}\) denote the working matrix, we can rewrite (\ref{partialdev}) as
\begin{equation} \label{partialdevwork}
\mS^c + \mS^r - 2\mW + \mG^{(q)} \odot \mA + \mLambda^{(q)} + (1/\rho)(\mTheta - \mGamma^{(q+1)}) =0.
\end{equation}

We only need to solve equation (\ref{partialdevwork}) to obtain the estimate of the precision matrix. We can solve (\ref{partialdevwork}) with the graphical Lasso algorithm \citep{friedman2008sparse}, i.e., updating one row or one column with the remaining ones fixed until all rows or columns are updated once. We provide the detailed calculation procedure for (\ref{partialdevwork}) is provided in Appendix \ref{mThetastepcompapp}. Performing the above calculation procedure for each column of the estimated matrix yields an estimate of the precision matrix, denoted as \(\widehat{\mTheta}\).

\subsection{Hyperparameters selection}
In this section, we discuss how to determine the hyperparameters, including the window width \(w\), the number of clusters \(K\), and the regularization parameter \(\lambda\).

\textbf{Window size \(w\).} Recall that we perform clustering on subsequences of length \(nw\) rather than individual observations. Assumptions about the block Toeplitz structure of the estimated precision matrix ensure that each cluster is time-invariant, allowing us to learn the cross-time correlation. However, a larger window size can result in pseudo-correlation, while a smaller window size can affect the quality of the subsequent imaging, both of which will impact the imaging dataset and thus compromise the forecasting. \citet{DBLP:conf/ijcai/HallacVBL18} experimentally demonstrate that the window size for a specific problem is relatively robust to the estimation of precision matrices and clustering, and they prefer a relatively small \(w\). In the empirical analysis in Section \ref{sec5}, we compare the effects of different window sizes on the final forecasting performance.

\textbf{The number of clusters \(K\).} The number of clusters determines the number of classes in the imaging dataset \(\mathcal{D}\). A larger \(K\) increases the complexity of the classification problem, which may lead to the neural network failing to obtain effective deep representations of the raw data, thus reducing the performance of forecasting. Since there is no prior information on the ITS dataset, we choose the model-independent criterion, Bayesian information criterion (BIC) \citep{schwarz1978estimating} , for determining the number of clusters. It is formally defined as:
\[
\begin{aligned}
\text{BIC} &= - 2(\text{maximized log likelihood}) + \log|P| \times \text{(No. of estimated parameters)},
\end{aligned}
\]
and the form of BIC for our problem is
\begin{equation} \label{BICcriterion}
\text{BIC} (K)= \sum_{k=1}^{K}-|P_k|\log\det\widehat{\mTheta}_{k} + |P_k|nw(1+\ln 2\pi) - \ln|P_k|((w-1)n^2 + n(n+1)/2), 
\end{equation}
where \(\widehat{\mTheta}_{k}\) is the estimated precision matrix corresponding to the \(k\)-th cluster. The number of clusters we choose is \(K^* = \argmin_K \text{BIC}(K)\).

\textbf{Regularization parameter \(\lambda\).} We utilize the popular \(V\)-fold cross-validation to choose the regularization parameter. Taking the \(k\)-th cluster as an example, all samples inside the cluster are divided into \(V\) disjoint subgroups, with the index of the \(v\)-th subgroup denoted by \(P_{k_v}\) for \(v=1,2,\cdots, V\), i.e., \(\bigcup^V_{v=1}P_{k_v} = P_k\). The \(V\)-fold cross-validation score is defined as:
\[ 
\text{CV}(\lambda_k) =\sum_{v=1}^{V}\left(|P_{k_v}|\log \det \widehat{\mTheta}^{-v}_k(\lambda_k) + \sum_{i\in P_{k_v}}(Y^l_i)^\top\widehat{\mTheta}^{-v}_k(\lambda_k)Y^l_i +(Y^u_i)^\top\widehat{\mTheta}^{-v}_k(\lambda_k)Y^u_i\right),   
\]
where \(\widehat{\mTheta}^{-v}_k(\lambda_k)\) denotes the estimated precision matrix of the \(k\)-th cluster using \(\lambda_k\) based on all samples in \(P_k\) except those in \(P_{k_v}\). Then, we choose \(\lambda^*_k = \arg\max_{\lambda_k} \text{CV}(\lambda_k)\) as the best regularization parameter for estimating the final precision matrix based on the entire samples in \(P_k\).

\section{Convergence of the algorithm} \label{sec4}
As discussed in Section \ref{sec3}, we can solve the optimization problem (\ref{optall}) in an alternating manner. First, we use the Viterbi algorithm to find the optimal assignment set with fixed precision matrices. Then, we estimate precision matrices using the samples within each assignment set. The Viterbi algorithm can find the globally optimal allocation set with a complexity of \(\mathcal{O}(KT)\). Therefore, we only need to analyze whether using the ADMM algorithm to solve optimization problem (\ref{Mstep}) can converge to the global optimum.

In this section, we prove that the sequence \((\mTheta^{(q)}, \mGamma^{(q)}, \mLambda^{(q)})\) produced by the ADMM algorithm converges to \((\widehat{\mTheta}^{+}, \widehat{\mGamma}^{+}, \widehat{\mLambda}^{+})\), where \((\widehat{\mTheta}^{+}, \widehat{\mGamma}^{+})\) is an optimal solution of \((\ref{Mstep})\), and \(\widehat{\mLambda}^{+}\) is the optimal dual variable. This automatically proves that Algorithm \ref{completeMstep} in Appendix \ref{mThetastepcompapp} provides an optimal solution to (\ref{Mstep}). Before presenting the theoretical results, we define some necessary notations for clarity. Let \(\mD\) be a \(2nw \times 2nw\) matrix defined as 
\[
\mD = \left[
\begin{array}{cc}
\rho \mI_{nw\times nw} & \bm{0}\\
\bm{0} & (1/\rho) \mI_{nw\times nw} 
\end{array}
\right].
\]

Define the norm \(\lVert \cdot \rVert^2_{\mD}\) as \(\lVert \mU \rVert^2_{\mD} = \langle \mU, \mD \mU \rangle\) and its corresponding inner product \(\langle \cdot, \cdot\rangle_{\mD}\) as \(\langle \mU, \mV \rangle_{\mD} = \langle \mU, \mD \mV \rangle\), where \(\mU\) and \(\mV\) are appropriate matrices. Before presenting the main theorem on the global convergence of Algorithm \ref{completeMstep}, we introduce the following auxiliary lemma. 

\begin{lemma} \label{lemma41}
Assume that (\(\widehat{\mTheta}^{+}, \widehat{\mGamma}^{+}\)) is an optimal solution of (\ref{Mstep}), and \(\widehat{\mLambda}^{+}\) is the corresponding optimal dual variable associate with the equality constrain \(\mTheta =\mGamma\). Then the sequence \(\left\{\left(\mTheta^{(q)}, \mGamma^{(q)}, \mLambda^{(q)}\right)\right\}\) produced by ADMM algorithm satisfies 
\begin{equation}
\lVert \mU^{(q)} - \mU^+\rVert^2_\mD -  \lVert \mU^{(q+1)} - \mU^+\rVert^2_\mD \geq \lVert \mU^{(q)} - \mU^{(q+1)}\rVert^2_\mD,
\end{equation}
where \(\mU^+ = \left(\widehat{\mTheta}^{+}, \widehat{\mGamma}^{+}\right)^\top\) and \(\mU^{(q)} = \left({\mTheta}^{(q)}, {\mGamma}^{(q)}\right)^\top\). 
\end{lemma}

\textbf{Remark 1.} An analogous conclusion was obtained by \citet{xue2012positive} in their study of Lasso-penalized estimation of high-dimensional covariance matrices. Their main objective was to ensure that the estimated large covariance matrix remains positive definite, which differs significantly from our goal of accurately estimating a block Toeplitz sparse precision matrix.

\textbf{Remark 2.} Our optimization objective is to jointly estimate the precision matrices for ITS using both the upper and lower bounds, which differs from the existing literature on estimating covariance and precision matrices for high-dimensional point-valued data.

\textbf{Remark 3.} The rationale for assuming that the structure of the precision matrix to be estimated exhibits block Toeplitz sparsity stems from the dependency characteristics of the subsequences used for clustering, as discussed in detail in Section \ref{sec21}. This assumption differs from the structural assumptions typically made about the covariance matrix for high-dimensional point-valued data.

Based on Lemma \ref{lemma41}, we derive the following main convergence result.
\begin{theorem}\label{theorem41}
The sequence \(\left\{\left(\mTheta^{(q)}, \mGamma^{(q)}, \mLambda^{(q)}\right)\right\}\) produced from any starting point converges to an optimal solution of (\ref{Mstep}), that is, 

(a) \(\lVert \mU^{(q)} - \mU^{(q+1)}\rVert_\mD \to 0\);

(b) \(\left\{\mU^{(q)}\right\}\) located in a compact region;

(c) \(\lVert \mU^{(q)} - \mU^{+}\rVert_\mD\) is monotonically non-increasing.
\end{theorem}

Theorem \ref{theorem41} (a) indicates that as \(q\) increases, the iterations of \(\mU^{(q)}\) converge to each other, demonstrating the convergence of the sequence.
Theorem \ref{theorem41} (b) asserts that \(\{\mU^{(q)}\}\) remains within a compact region, ensuring the sequence does not grow unboundedly but stays within a finite, well-defined area. In Theorem \ref{theorem41} (c), monotonically non-increasing means that each subsequent iteration \(q+1\) either reduces or maintains the same distance to \(\mU^{+}\) compared to iteration \(q\). This property guarantees that the sequence \(\{\mU^{(q)}\}\) consistently progresses towards the optimal solution \(\mU^{+}\) as defined by the optimization problem (\ref{Mstep}).

\section{Empirical analysis} \label{sec5}

In this section, we apply the proposed method to the analysis of stock market data, and discuss the impact of the number of clusters, window size, and different models on prediction performance. Before modeling, we first present the following experimental settings.

We set the collection of cluster numbers as \(\gK = \{2, 3, 4, 5, 6, 7\}\), and select the optimal number of clusters from this set based on the BIC shown in (\ref{BICcriterion}). It is evident that as \(K\) decreases (due to the presence of multiple complex patterns in financial data, which can lead to less accurate clustering), the resulting image dataset contains fewer categories. Consequently, the accuracy of classification by deep learning models increases, leading to overfitting. Conversely, as \( K \) increases, the image dataset contains more categories, making the classification task more complex. This results in insufficient feature extraction, leading to underfitting. It is important to note that the \(K\) selected based on BIC is optimal for clustering tasks, not for prediction tasks. We set the collection of possible values for the window width to \(\gW = \{10, 15, 20, 25, 30, 40\}\). It is worth noting that, for better utilization of the feature extraction network, the size of the window width is consistent with the length of the information set used during prediction.

We compare our proposed method with both classical statistical learning and deep learning methods. The statistical learning methods include linear models such as support vector machines (SVM), elastic net regression (ENR), and Bayesian ridge regression (BRR), as well as ensemble learning models like decision trees (DT), AdaBoost, gradient boosting (GB), and random forests (RF), in addition to the nearest neighbors (NN) algorithm. The deep learning methods include 26 approaches (can be found in Table \ref{deeplearningmetho}.) categorized into six major types: multilayer perceptrons (MLP), recurrent neural networks (RNNs), convolutional neural networks (CNNs), Transformers, wavelet-based models, and hybrid models. These six types have achieved state-of-the-art performance in point value time series modeling \citep{ismail2019deep, blazquez2021review, lines2018time, lim2021time}. Considering that these methods are only applicable to point-valued time series, we first model the center and range of the ITS, and then reconstruct the interval (i.e., the upper and lower bounds of the interval). For the proposed method, we also first obtain the deep representations corresponding to the original ITS, and then convert them into the center and range for modeling. It is worth noting that we only combine the proposed feature extraction process with classical statistical learning methods to verify whether the proposed method can improve the predictive power of the models.

To compare the performance of different methods, we employ multiple forecast criteria. The first criterion we consider is the mean distance error (MDE), which is defined as follows:
\[
\text{MDE}_d = \frac{\sum_{t = T_e + 1}^{T_e+T_f}d(y_t, \widehat{y}_t)}{T_f},
\]
where \((y_t)^{T_e + T_f}_{t = T_e + 1}\) represents the actual values and \((\widehat{y}_t)^{T_e + T_f}_{t = T_e + 1}\) represents the predicted values. \(T_e + 1\) and \(T_f\) are the start and end points of the prediction, respectively. The function \(d(\cdot,\cdot)\) is an appropriate distance metric between paired intervals. We consider two choices for \(d(\cdot,\cdot)\), namely 
\[
d_1(y_t, \widehat{y}_t)  = \left((y^c_{t} - \widehat{y}^c_t)^2 + (y^r_{t} - \widehat{y}^r_t)^2\right)^{1/2}, \ d_2(y_t, \widehat{y}_t) = D_K(y_t, \widehat{y}_t),
\]
where \(y^c_{t} = (y^l_{t}+ y^u_{t}) /2\) and \(y^r_{t} = (y^u_{t}- y^l_{t}) /2\) represent the upper and lower bounds of the interval \(y_t\), respectively, \(D_K\) is an appropriate distance metric between paired intervals, and interested readers can refer to \citet{han2016vector} for more details. Following \citet{sun2018threshold}, we select the kernel function \( \left[
\begin{array}{cc}	
5 & 1\\
1 & 1\\
\end{array}
\right]\) for the \(D_K\)-distance.

The data used for our modeling consists of stock data from the 500 companies in the Standard\&Poor's 500 (S\&P 500) index. These 500 companies are spread across 10 sectors: Basic Materials (21 companies), Communication Services (26 companies), Consumer Cyclical (58 companies), Consumer Defensive (36 companies), Energy (22 companies), Financial Services (69 companies), Healthcare (65 companies), Industrials (73 companies), Technology (71 companies), Utilities (30 companies). Due to our own constraints, we were only able to obtain daily stock information for 81 out of the 500 publicly traded companies from September 5, 2012, to September 1, 2017. This data includes seven indicators: Date, Open, High, Low, Close, Adj Close, and Volume. The basic information of these 81 companies is provided in Appendix \ref{informationsselected}. We use the daily high and low stock prices as the upper and lower bounds of the interval, respectively, to construct the ITS. Consequently, we obtain an ITS with a length of \(T = 1823\) and a dimension of \(n = 81\). We determine the final structure of the feature extraction network based on training loss and test accuracy. The selected network structures show the classification performance on the stock dataset in Table \ref{selectnetstock}. 

\begin{table}[H]
\setlength{\abovecaptionskip}{0pt}%
\setlength{\belowcaptionskip}{3pt}%
\centering
\footnotesize
\caption{Prediction results (\(\text{MDE}_{d_1}/\text{MDE}_{d_2}\)) of classical statistical learning.} \label{classicalstock}
\setlength{\tabcolsep}{0.6mm}{
\begin{tabular}{c|c|c|c|c|c|c}
\hline
\multirow{2}{*}{Method} &\multicolumn{6}{c}{\(w\)} \\
\cline{2-7}
& 10 & 15 & 20 & 25 & 30 & 40\\
\hline
SVM &0.0504/0.1070&0.0503/0.1065&0.0506/0.1084&0.0499/0.1059&0.0495/0.1058&0.0491/0.1040\\
\hline
ENR &0.0423/0.0857&0.0424/0.0858&0.0431/0.0873&0.0424/0.0859&0.0426/0.0862&0.0420/0.0850\\
\hline
BRR &0.0070/0.0162&0.0070/0.0161&0.0071/0.0165&0.0070/0.0162&0.0069/0.0159&0.0071/0.0164\\
\hline
DT &0.0103/0.0228&0.0101/0.0223&0.0105/0.0233&0.0103/0.0228&0.0103/0.0227&0.0104/0.0229\\
\hline
AdaBoost &0.0097/0.0221&0.0095/0.0219&0.0094/0.0219&0.0095/0.0220&0.0091/0.0210&0.0091/0.0210\\
\hline
GB &0.0073/0.0168&0.0073/0.0166&0.0074/0.0172&0.0073/0.0169&0.0073/0.0166&0.0075/0.0170\\
\hline
RF &0.0073/0.0167&0.0071/0.0164&0.0073/0.0169&0.0072/0.0166&0.0071/0.0163&0.0072/0.0166\\
\hline
NN &0.0079/0.0178&0.0078/0.0177&0.0080/0.0181&0.0079/0.0181&0.0079/0.0178&0.0081/0.0183\\
\hline
\end{tabular}
}
\end{table}

Table \ref{classicalstock} presents the prediction results of the selected statistical learning methods on the stock dataset across different window sizes. We compare the methods from both horizontal and vertical perspectives. Horizontally, with the method fixed, the window size seems to have a limited impact on model performance. For example, the performance metrics of SVM across the six window sizes are 0.0504/0.1070, 0.0503/0.1065, 0.0506/0.1084, 0.0499/0.1059, 0.0495/0.1058, and 0.0491/0.1040, showing no significant change with varying window sizes. This suggests that even the smallest window size sufficiently captures the necessary predictive order. Vertically, with the window size fixed, ensemble learning methods exhibit similar performance, outperforming DT and NN, and significantly outperforming the linear models SVM and ENR. For instance, when the window size is \(w=10\), the performance metrics for the eight methods are 0.0504/0.1070, 0.0423/0.0857, 0.0070/0.0162, 0.0103/0.0228, 0.0097/0.0221, 0.0073/0.0168, 0.0073/0.0167, and 0.0079/0.0178. Overall, AdaBoost, GB, and RF demonstrate similar performance and reach optimal levels, outperforming DT and significantly outperforming SVM and ENR.

\begin{table}[H]
\setlength{\abovecaptionskip}{0pt}%
\setlength{\belowcaptionskip}{3pt}%
\centering
\caption{Prediction results (\(\text{MDE}_{d_1}/\text{MDE}_{d_2}\)) of deep learning methods.} \label{deeplearningmetho}
\setlength{\tabcolsep}{0.2mm}{
\scriptsize
\begin{tabular}{c|l|c|c|c|c|c|c}
\hline
\multirow{2}{*}{Types} & \multirow{2}{*}{Method} &\multicolumn{6}{c}{\(w\)} \\
\cline{3-8}
& & 10 & 15 & 20 & 25 & 30 & 40\\
\hline
\multirow{2}{*}{MLP} & MLP &0.0225/0.0456&0.0200/0.0411&0.0185/0.0380&0.0176/0.0367&0.0171/0.0352&0.0149/0.0312\\
& gMLP &0.0090/0.0200&0.0087/0.0195&0.0089/0.0198&0.0091/0.0204&0.0090/0.0199&0.0090/0.0200\\
\hline

\multirow{3}{*}{RNNs} & RNNPlus &0.0094/0.0208&0.0095/0.0207&0.0094/0.0211&0.0096/0.0213&0.0094/0.0207&0.0094/0.0208\\
& RNNAttention &0.0352/0.0748&0.0372/0.0866&0.0361/0.0832&0.0415/0.0898&0.0438/0.0979&0.0413/0.0935\\
& TSSequencerPlus &0.0110/0.0236&0.0110/0.0237&0.0110/0.0238&0.0111/0.0241&0.0114/0.0243&0.0112/0.0243\\
\hline

\multirow{14}{*}{CNNs} & FCN &0.0229/0.0539&0.0211/0.0480&0.0185/0.0415&0.0178/0.0392&0.0170/0.0370&0.0171/0.0367\\
& FCNPlus &0.0228/0.0536&0.0193/0.0443&0.0191/0.0425&0.0182/0.0401&0.0169/0.0370&0.0163/0.0353\\
& ResNet &0.0472/0.1171&0.0435/0.1055&0.0373/0.0879&0.0334/0.0776&0.0310/0.0706&0.0276/0.0617\\
& ResNetPlus &0.0461/0.1154&0.0428/0.1040&0.0381/0.0894&0.0340/0.0790&0.0302/0.0691&0.0279/0.0626\\
& XResNet1dPlus &0.0566/0.1247&0.0560/0.1246&0.0575/0.1278&0.0595/0.1310&0.0586/0.1291&0.0620/0.1329\\
& ResCNN &0.0205/0.0485&0.0197/0.0447&0.0181/0.0405&0.0172/0.0377&0.0162/0.0355&0.0154/0.0336\\
& TCN &0.0131/0.0281&0.0116/0.0252&0.0100/0.0221&0.0094/0.0211&0.0092/0.0204&0.0101/0.0223\\
& InceptionTime &0.0320/0.0769&0.0335/0.0811&0.0316/0.0756&0.0311/0.0727&0.0290/0.0673&0.0268/0.0610\\
& InceptionTimePlus &0.0321/0.0757&0.0321/0.0756&0.0306/0.0723&0.0300/0.0692&0.0284/0.0648&0.0287/0.0635\\
& XceptionTime &0.0670/0.1860&0.0545/0.1502&0.0490/0.1339&0.0475/0.1215&0.0438/0.1126&0.0417/0.1083\\
& XceptionTimePlus &0.0675/0.1865&0.0547/0.1496&0.0502/0.1338&0.0410/0.1079&0.0427/0.1103&0.0476/0.1230\\
& OmniScaleCNN &0.0314/0.0642&0.0276/0.0564&0.0252/0.0519&0.0251/0.0515&0.0233/0.0482&0.0222/0.0456\\
& XCM &0.0173/0.0424&0.0171/0.0414&0.0172/0.0414&0.0165/0.0393&0.0159/0.0376&0.0150/0.0352\\
& XCMPlus &0.0182/0.0449&0.0176/0.0425&0.0170/0.0409&0.0160/0.0385&0.0161/0.0379&0.0155/0.0360\\
\hline

\multirow{4}{*}{Transformers} & TransformerModel &0.0442/0.1046&0.0498/0.1144&0.0518/0.1245&0.0499/0.1222&0.0626/0.1408&0.0682/0.1739\\
& TST &0.0563/0.1260&0.0576/0.1236&0.0560/0.1202&0.0572/0.1257&0.0587/0.1307&0.0555/0.1226\\
& TSTPlus &0.0292/0.0593&0.0345/0.0702&0.0292/0.0596&0.0335/0.0681&0.0321/0.0653&0.0341/0.0693\\
& TSiT &0.0090/0.0199&0.0088/0.0195&0.0090/0.0201&0.0092/0.0203&0.0085/0.0191&0.0088/0.0195\\
\hline

\multirow{1}{*}{Wavelet} & mWDN &0.0127/0.0269&0.0127/0.0269&0.0126/0.0266&0.0144/0.0300&0.0132/0.0278&0.0130/0.0272\\
\hline

\multirow{2}{*}{Hybrid} & RNN\_FCN &0.0183/0.0431&0.0172/0.0392&0.0163/0.0366&0.0158/0.0349&0.0144/0.0319&0.0150/0.0328\\
& RNN\_FCNPlus &0.0187/0.0437&0.0167/0.0384&0.0164/0.0366&0.0154/0.0341&0.0151/0.0331&0.0143/0.0311\\
\hline

\end{tabular}
}
\end{table}

Table \ref{deeplearningmetho} presents the prediction results of the selected deep learning methods for stock datasets across different window sizes. The table highlights significant variations in prediction performance between different types of deep learning methods. For instance, the Hybrid method consistently outperforms most CNN and Transformer-based methods across all window sizes. Notably, there are also substantial differences within the same type of method. For example, gMLP, a variant of the MLP, significantly outperforms the traditional MLP, and RNNPlus shows superior performance compared to RNNAttention.

We compared these methods from both horizontal and vertical perspectives. Horizontally, the effect of window size on model performance appears to be minimal when the method is fixed. For instance, the performance metrics of gMLP across six window sizes are 0.0090/0.0200, 0.0087/0.0195, 0.0089/0.0198, 0.0091/0.0204, 0.0090/0.0199, and 0.0090/0.0200, showing no significant differences. This aligns with previous findings that even the smallest window sizes capture the necessary information for prediction. Vertically, when fixing the window size, models such as gMLP, RNNPlus, TSSequencerPlus, TCN, TSiT, and mWDN show similar performance, outperforming Hybrid models like RNN\_FCN and RNN\_FCNPlus, as well as CNNs type models like XCM, XCMPlus, and ResCNN. These models also significantly outperform others such as ResNet, XceptionTime, and TST. For example, when the window size is 10, the performance metrics for these methods are as follows: 0.0090/0.0200, 0.0094/0.0208, 0.0110/0.0236, 0.0131/0.0281, 0.0090/0.0199, 0.0127/0.0269, 0.0183/0.0431, 0.0187/0.0437, 0.0173/0.0424, 0.01\\82/0.0449, 0.0205/0.0485, 0.0472/0.1171, 0.0670/0.1860, and 0.0563/0.1260.

\begin{figure}[H]
\centering 
\begin{minipage}[t]{0.96\textwidth}
\centering
\includegraphics[scale=0.32]{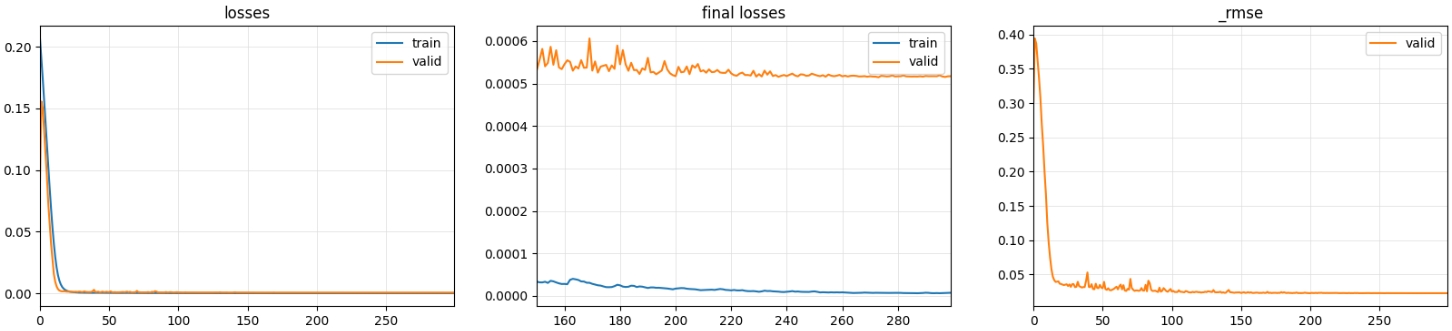}
\end{minipage}

\begin{minipage}[t]{0.96\textwidth}
	\centering
	\includegraphics[scale=0.32]{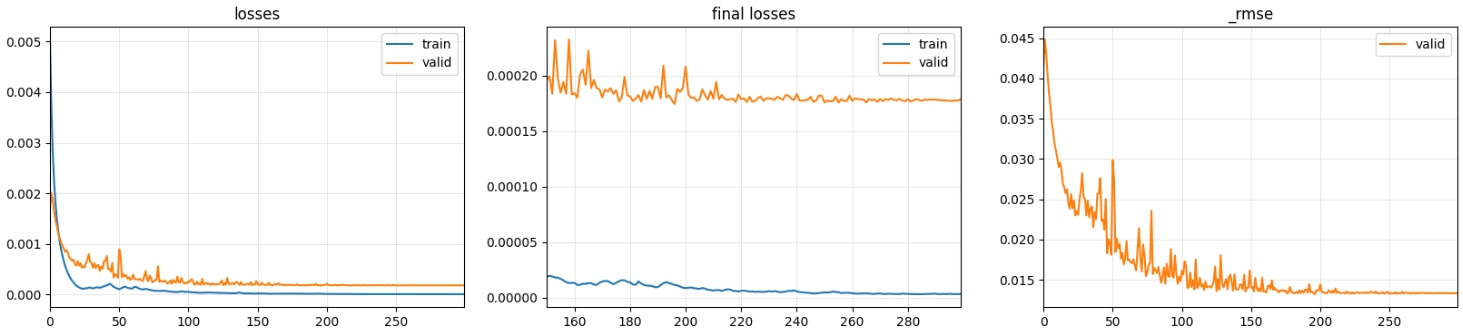}
\end{minipage}

\begin{minipage}[t]{0.96\textwidth}
	\centering
	\includegraphics[scale=0.32]{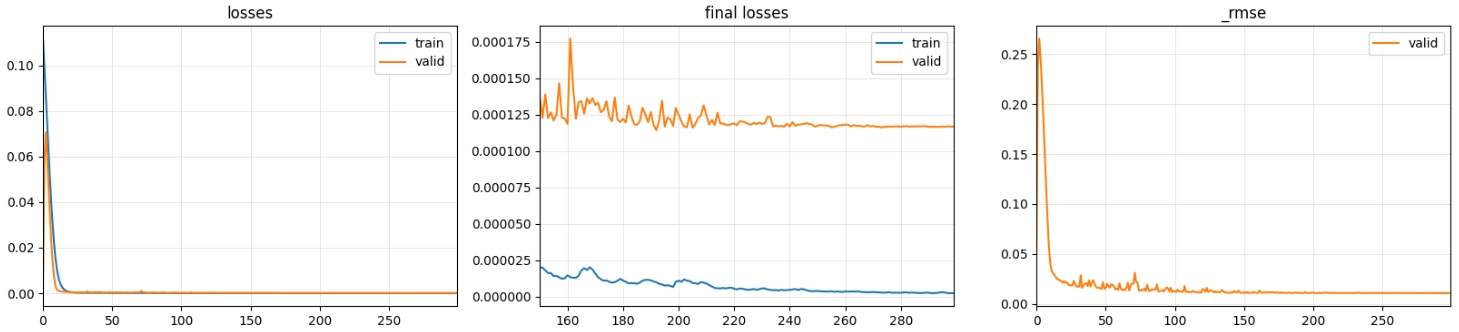}
\end{minipage}

\begin{minipage}[t]{0.96\textwidth}
	\centering
	\includegraphics[scale=0.32]{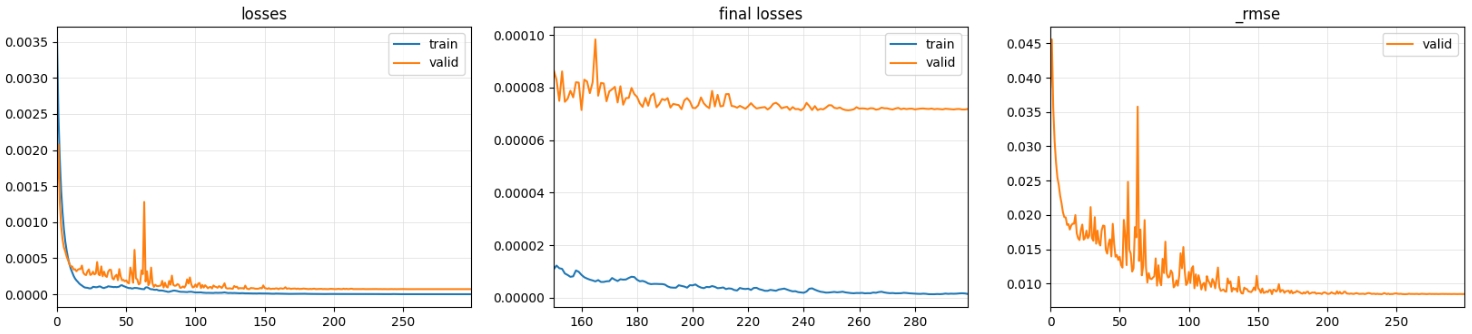}
\end{minipage}

\begin{minipage}[t]{0.96\textwidth}
	\centering
	\includegraphics[scale=0.32]{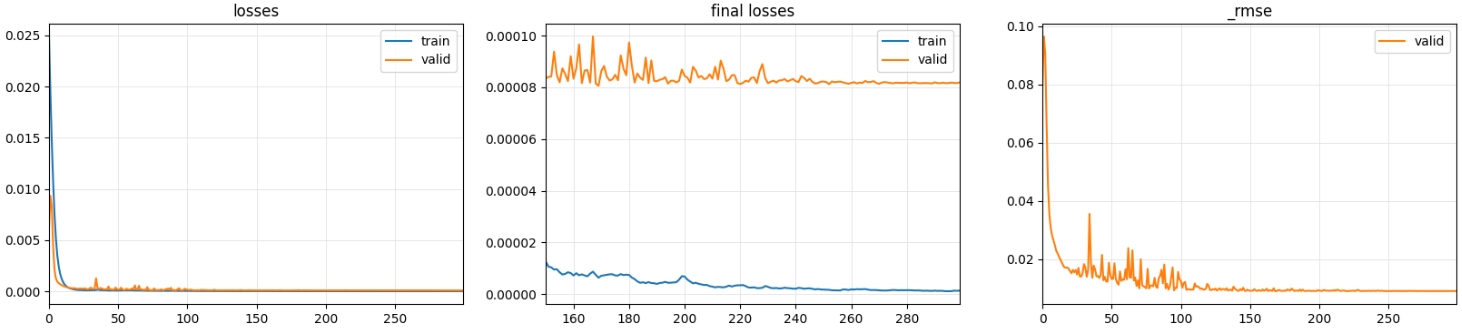}
\end{minipage}

\begin{minipage}[t]{0.96\textwidth}
	\centering
	\includegraphics[scale=0.32]{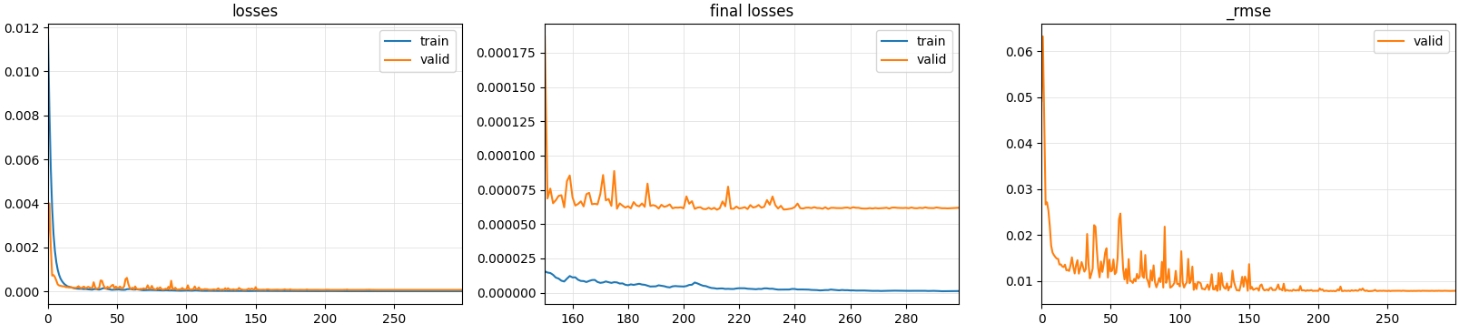}
\end{minipage}

\caption{From top to bottom, the figures show the loss on the training set and the root mean squared error on the test set during the training process of the deep learning method ResCNN for window widths of 10, 15, 20, 25, 30, and 40, respectively.}\label{convergencefigure}
\end{figure}

Recall that the essence of our proposed method is to construct a feature extraction network to enhance the model's predictive performance. Before constructing the feature extraction network, we should first determine an appropriate network structure. Considering that there are many models for image classification tasks, we limit our selection to well-known ResNet \citep{DBLP:conf/cvpr/HeZRS16} and VGG \citep{DBLP:journals/corr/SimonyanZ14a} models with different layers, as well as a fine-grained image classification network WS-DAN \citep{hu2019see}. The specific networks are shown in Table \ref{selectnetstock}. The reason for considering the fine-grained image classification network is that in the image dataset \(\mathcal{D}\), images of different classes have high similarity, as shown in Figure \ref{technology_roadmap}.

\begin{table}[H]
\small
\setlength{\abovecaptionskip}{0pt}%
\setlength{\belowcaptionskip}{3pt}%
\centering
\caption{Classification performance of different network structures on the image dataset corresponding to stock data under different window widths and numbers of clusters.} \label{selectnetstock}
\setlength{\tabcolsep}{0.2mm}{
\begin{tabular}{c|c|c|c|c|c|c|c|c|c|c|c}
\hline
\multirow{2}{*}{\(K\)} & \multirow{2}{*}{\(w\)} & \multicolumn{4}{c|}{VGG} & \multicolumn{5}{c|}{ResNet} &  \multirow{2}{*}{WS-DAN}\\
\cline{3-11}
& &  11 & 13 & 16 & 19 & 18 & 34 & 50 & 101 & 152 & \\
\hline

\multirow{6}{*}{2} & 10 &0.70/50.4&0.70/50.2&0.70/50.4&0.70/50.4&0.71/50.4&0.71/50.4&0.70/50.4&0.70/50.4&0.70/50.4&0.20/90.2\\
& 15 &0.70/50.9&0.70/50.9&0.70/50.9&0.70/50.9&0.70/50.9&0.70/50.9&0.70/50.9&0.70/50.9&0.70/50.9&0.21/91.2\\
& 20 &0.70/52.7&0.70/52.7&0.70/52.7&0.70/52.7&0.70/52.7&0.70/52.7&0.70/52.7&0.70/52.7&0.70/52.7&0.20/90.1\\
& 25 &0.70/52.9&0.70/52.9&0.70/52.9&0.70/52.9&0.70/52.9&0.70/52.9&0.70/52.9&0.70/52.9&0.70/52.9&0.23/89.8\\
& 30 &0.70/50.9&0.70/53.2&0.70/53.2&0.70/53.2&0.70/53.2&0.70/53.2&0.70/53.2&0.70/53.2&0.70/53.2&0.20/92.1\\
& 40 &0.70/53.5&0.70/53.5&0.70/53.5&0.70/53.5&0.71/53.5&0.71/53.5&0.71/53.5&0.71/53.5&0.71/53.5&0.20/91.6\\
\hline

\multirow{6}{*}{3} & 10 &1.10/37.4&1.11/37.9&1.10/37.4&1.11/37.4&1.11/37.9&1.11/37.9&1.11/37.9&1.11/37.9&1.11/32.2&0.45/83.1\\
& 15 &1.11/40.6&1.11/40.6&1.10/40.6&1.10/40.1&1.11/40.6&1.12/40.6&1.10/40.6&1.11/40.6&1.11/40.6&0.44/82.8\\
& 20 &1.11/42.0&1.11/42.0&1.11/42.0&1.11/42.0&1.11/42.0&1.11/42.0&1.11/42.0&1.11/42.0&1.11/42.0&0.43/84.1\\
& 25 &1.11/43.5&1.11/43.5&1.11/43.5&1.11/43.5&1.11/43.5&1.11/43.5&1.11/43.5&1.11/43.5&1.11/43.5&0.44/83.3\\
& 30 &1.10/45.0&1.10/45.0&1.10/45.0&1.10/45.0&1.10/45.0&1.10/45.0&1.10/45.0&1.10/45.0&1.10/45.0&0.49/85.1\\
& 40 &1.10/46.5&1.10/46.5&1.10/46.5&1.10/46.5&1.10/46.5&1.10/46.5&1.10/46.5&1.10/46.5&1.10/46.5&0.41/85.1\\
\hline

\multirow{6}{*}{4} & 10 & 1.39/26.6 & 1.39/26.6 & 1.39/27.1 & 1.39/27.1 & 1.40/26.7 & 1.40/26.6 & 1.39/27.1 & 1.39/27.1 & 1.40/27.1 & 0.54/78.9\\
& 15 & 1.40/26.5 & 1.40/26.2 & 1.39/26.5 & 1.40/26.5 & 1.40/26.5 & 1.41/26.5 & 1.40/26.5 & 1.40/26.5 & 1.40/26.5 & 0.55/79.2\\
& 20 & 1.40/27.8 & 1.40/27.8 & 1.40/27.8 & 1.40/27.8 & 1.40/27.8 & 1.40/27.8 & 1.40/27.8 & 1.40/27.8 & 1.40/27.8 & 0.57/80.1\\
& 25 & 1.40/28.4 & 1.40/28.4 & 1.40/28.4 & 1.40/28.4 & 1.40/28.4 & 1.40/28.4 & 1.40/28.4 & 1.40/28.4 & 1.40/28.4 & 0.56/81.3\\
& 30 & 1.40/29.0 & 1.40/29.0 & 1.40/29.0 & 1.40/29.0 & 1.40/29.0 & 1.40/29.0 & 1.40/29.0 & 1.40/29.0 & 1.40/29.0 & 0.58/82.4\\
& 40 & 1.40/29.6 & 1.40/29.6 & 1.40/29.6 & 1.40/29.6 & 1.40/29.6 & 1.40/29.6 & 1.40/29.6 & 1.40/29.6 & 1.40/29.6 & 0.59/83.2\\
\hline
\multirow{6}{*}{5} & 10 & 1.59/25.6 & 1.59/25.6 & 1.59/25.6 & 1.60/25.6 & 1.62/26.1 & 1.64/26.1 & 1.60/25.6 & 1.60/25.6 & 1.60/25.6 & 0.63/76.2\\
& 15 & 1.61/24.6 & 1.61/24.6 & 1.61/24.6 & 1.61/24.6 & 1.62/24.6 & 1.62/24.6 & 1.61/24.6 & 1.61/24.6 & 1.61/24.6 & 0.62/77.4\\
& 20 & 1.62/26.0 & 1.62/26.0 & 1.62/26.0 & 1.62/26.0 & 1.62/26.0 & 1.62/26.0 & 1.62/26.0 & 1.62/26.0 & 1.62/26.0 & 0.64/78.1\\
& 25 & 1.62/27.0 & 1.62/27.0 & 1.62/27.0 & 1.62/27.0 & 1.62/27.0 & 1.62/27.0 & 1.62/27.0 & 1.62/27.0 & 1.62/27.0 & 0.65/79.3\\
& 30 & 1.62/28.0 & 1.62/28.0 & 1.62/28.0 & 1.62/28.0 & 1.62/28.0 & 1.62/28.0 & 1.62/28.0 & 1.62/28.0 & 1.62/28.0 & 0.66/80.2\\
& 40 & 1.62/29.0 & 1.62/29.0 & 1.62/29.0 & 1.62/29.0 & 1.62/29.0 & 1.62/29.0 & 1.62/29.0 & 1.62/29.0 & 1.62/29.0 & 0.67/81.4\\
\hline

\multirow{6}{*}{6} 
& 10 &1.79/21.8&1.79/21.8&1.79/21.8&1.79/21.8&1.80/22.4&1.81/22.4&1.80/22.4&1.80/22.4&1.80/22.4&0.71/71.7\\
& 15 &1.79/22.4&1.79/22.4&1.79/22.4&1.79/22.4&1.80/18.9&1.80/22.9&1.79/22.9&1.79/22.9&1.79/22.9&0.71/72.6\\
& 20 &1.80/23.5&1.80/23.5&1.80/23.5&1.80/23.5&1.80/23.5&1.80/23.5&1.80/23.5&1.80/23.5&1.80/23.5&0.68/73.1\\
& 25 &1.80/24.0&1.80/24.0&1.80/24.0&1.80/24.0&1.80/24.0&1.80/24.0&1.80/24.0&1.80/24.0&1.80/24.0&0.73/70.9\\
& 30 &1.80/24.5&1.80/24.5&1.80/24.5&1.80/24.5&1.80/24.5&1.80/24.5&1.80/24.5&1.80/24.5&1.80/24.5&0.69/72.8\\
& 40 &1.80/25.0&1.80/25.0&1.80/25.0&1.80/25.0&1.80/25.0&1.80/25.0&1.80/25.0&1.80/25.0&1.80/25.0&0.73/73.0\\
\hline

\multirow{6}{*}{7} 
& 10 &1.94/20.7&1.94/20.7&1.94/20.7&1.94/20.7&1.95/20.7&1.96/21.3&1.94/21.3&1.95/21.3&1.95/21.3&0.81/68.9\\
& 15 &1.93/19.9&1.93/19.9&1.93/19.9&1.93/19.9&1.94/19.9&1.95/19.9&1.93/19.9&1.93/19.9&1.93/19.9&0.81/68.9\\
& 20 &1.95/21.0&1.95/21.0&1.95/21.0&1.95/21.0&1.95/21.0&1.95/21.0&1.95/21.0&1.95/21.0&1.95/21.0&0.86/68.1\\
& 25 &1.95/21.5&1.95/21.5&1.95/21.5&1.95/21.5&1.95/21.5&1.95/21.5&1.95/21.5&1.95/21.5&1.95/21.5&0.81/70.9\\
& 30 &1.95/22.0&1.95/22.0&1.95/22.0&1.95/22.0&1.95/22.0&1.95/22.0&1.95/22.0&1.95/22.0&1.95/22.0&0.84/64.9\\
& 40 &1.95/22.5&1.95/22.5&1.95/22.5&1.95/22.5&1.95/22.5&1.95/22.5&1.95/22.5&1.95/22.5&1.95/22.5&0.88/65.7\\
\hline
\end{tabular}
}
\end{table}

From Table \ref{selectnetstock}, we can observe that for VGG, ResNet, and their variants, the training loss and test accuracy appear to be influenced primarily by the number of clusters rather than the window width. For instance, with the VGG-11 architecture and two clusters, the classification performance metrics across different window widths are 0.70/50.4, 0.70/50.9, 0.70/52.7, 0.70/52.9, 0.70/50.9, and 0.70/53.5, showing no significant variation. This suggests that for image datasets with high intra-class and inter-class similarity, the discriminative ability of network structures like VGG and ResNet is almost lost. When the window width is fixed, the performance of VGG and ResNet networks significantly declines as the number of clusters increases. For example, with a window width of 10, the performance of VGG-11 as the number of clusters increases is 0.70/50.4, 1.10/37.4, 1.39/26.6, 1.59/25.6, 1.79/21.8, and 1.94/20.7. The increase in the number of clusters leads to a substantial rise in classification complexity, thereby degrading the model’s performance. In contrast, WS-DAN demonstrates strong performance, with classification accuracy exceeding 80\% when the number of clusters is small (e.g., 2, 3, 4, 5), and still remaining above 65\% when the number of clusters is 7. Based on these observations, we select the fine-grained network WS-DAN as the final structure for the feature extraction network.

Based on the previously established feature extraction network WS-DAN, we use it to extract features from the image dataset and combine them with classical statistical learning methods for prediction. The corresponding results are shown in Table \ref{statisticalstock} below.

\begin{table}[H]
\setlength{\abovecaptionskip}{0pt}%
\setlength{\belowcaptionskip}{3pt}%
\centering
\caption{Prediction results (\(\text{MDE}_{d_1}/\text{MDE}_{d_2}\)) of statistical machine learning methods combination with the proposed feature extraction process.} \label{statisticalstock}
\setlength{\tabcolsep}{0.6mm}{
\begin{tabular}{c|c|c|c|c|c|c}
\hline
\multirow{2}{*}{Method} &\multicolumn{6}{c}{\(w\)} \\
\cline{2-7}
& 10 & 15 & 20 & 25 & 30 & 40\\
\hline
SVM &0.0386/0.0835&0.0402/0.0840&0.0374/0.0786&0.0373/0.0819&0.0368/0.0800&0.0377/0.0815\\
\hline
ENR &0.0318/0.0659&0.0325/0.0651&0.0314/0.0672&0.0328/0.0635&0.0340/0.0676&0.0308/0.0612\\
\hline
BRR &0.0054/0.0122&0.0052/0.0117&0.0053/0.0119&0.0053/0.0123&0.0054/0.0117&0.0054/0.0121\\
\hline
DT &0.0079/0.0182&0.0080/0.0167&0.0079/0.0170&0.0081/0.0170&0.0074/0.0175&0.0077/0.0179\\
\hline
AdaBoost &0.0073/0.0173&0.0075/0.0172&0.0070/0.0168&0.0075/0.0159&0.0073/0.0153&0.0069/0.0163\\
\hline
GB &0.0058/0.0127&0.0057/0.0130&0.0056/0.0132&0.0057/0.0123&0.0053/0.0127&0.0057/0.0132\\
\hline
RF &0.0055/0.0133&0.0054/0.0127&0.0053/0.0131&0.0057/0.0123&0.0053/0.0121&0.0052/0.0125\\
\hline
NN &0.0057/0.0140&0.0057/0.0128&0.0059/0.0142&0.0058/0.0133&0.0060/0.0142&0.0063/0.0140\\
\hline
\end{tabular}
}
\end{table}

In Table \ref{statisticalstock}, we observe similar patterns to those in Table \ref{classicalstock}. For example, the window width seems to have minimal impact on the performance of each method. For instance, with the AdaBoost model, the two performance metrics across different window widths are 0.0073/0.0173, 0.0075/0.0172, 0.0070/0.0168, 0.0075/0.0159, 0.0073/0.0153, and 0.0069/0.0163, showing no significant variation with changes in window width. Additionally, as expected, ensemble models outperform nearest neighbors and linear models. For example, with a window width of 10, the performance metrics for RF, GB, ENR, and NN are 0.0055/0.0133, 0.0058/0.0127, 0.0318/0.0659, and 0.0057/0.0140, respectively.

Comparing Table \ref{classicalstock} and Table \ref{statisticalstock}, we observe that regardless of the model's performance on the raw data, the prediction performance of the model significantly improves after feature extraction. For example, with a window width of 10, the performance metrics for SVM before and after feature extraction are 0.0504/0.1070 and 0.0386/0.0835, respectively; for AdaBoost, the metrics are 0.0097/0.0221 and 0.0073/0.0173. These results indicate that the proposed feature extraction process effectively captures meaningful representations of the original time series that enhance predictive performance. Comparing Table \ref{deeplearningmetho} and Table \ref{statisticalstock}, we observe that the Transformer-type model TSiT achieved the best prediction performance at a window size of 30, with the two performance metrics being 0.0085/0.0191. However, apart from SVM and ENR, the proposed feature extraction process combined with the remaining six models outperformed TSiT. For instance, at a window size of 10, AdaBoost achieved performance metrics of 0.0073/0.0173, both of which are better than those of TSiT.

The conclusions drawn from the comparison of these tables demonstrate that the proposed feature extraction process not only enhances the performance of classical statistical learning methods but also outperforms the current state-of-the-art deep learning methods. This validates the effectiveness of the proposed approach in extracting deep representations from large-scale interval-valued time series.

\section{Conclusions} \label{sec6}
In this paper, we proposed a representation learning method for large-scale interval-valued time series, which effectively enhanced the predictive performance of general statistical learning methods. The method was based on the automatic segmentation and clustering of interval-valued time series, which we formulated as a combinatorial optimization problem and provided an efficient solution using the majorization-minimization algorithm. Furthermore, we proved the convergence of the block Toeplitz sparse precision matrix estimation at the optimization level. Experimental results on large-scale stock data demonstrated that the proposed feature extraction method, combined with classical statistical approaches, outperformed the current state-of-the-art deep learning methods.

We identified two major challenges in the proposed feature extraction process. The first challenge was that the currently available imaging method for multivariate time series,  JRPs, resulted in high intra-class and inter-class similarity within the generated imaging dataset. Conventional CNNs, such as VGG and ResNet, struggled to extract features from these images, necessitating the use of more complex fine-grained networks. The second challenge arose when the window size was large, leading to a high-dimensional block Toeplitz precision matrix, which incurred substantial computational costs. To address these challenges, we actively explored new imaging methods for multivariate time series and efficient parallel optimization algorithms to further enhance the applicability of the proposed approach.

\section*{Acknowledgments}
\addcontentsline{toc}{section}{Acknowledgments}
The research work described in this paper was supported by the National Natural Science Foundation of China (Nos. 72071008).

\newpage 
\begin{appendix}
\section{Proofs for Results}\label{appendix}

{\bf Lemma 4.1} {\it 
Assume that (\(\widehat{\mTheta}^{+}, \widehat{\mGamma}^{+}\)) is an optimal solution of (\ref{Mstep}), and \(\widehat{\mLambda}^{+}\) is the corresponding optimal dual variable associate with the equality constrain \(\mTheta = \mGamma\). Then the sequence \(\left\{\left(\mTheta^{(q)}, \mGamma^{(q)}, \mLambda^{(q)}\right)\right\}\) produced by ADMM algorithm satisfies 
\begin{equation}
	\lVert \mU^{(q)} - \mU^+\rVert^2_\mD -  \lVert \mU^{(q+1)} - \mU^+\rVert^2_\mD \geq \lVert \mU^{(q)} - \mU^{(q+1)}\rVert^2_\mD,
\end{equation}
where \(\mU^+ = \left(\widehat{\mTheta}^{+}, \widehat{\mLambda}^{+}\right)^\top\) and \(\mU^{(q)} = \left({\mTheta}^{(q)}, {\mLambda}^{(q)}\right)^\top\). 
}
\begin{proof}
Since (\(\widehat{\mTheta}^{+}, \widehat{\mGamma}^{+}\)) is an optimal solution of (\ref{Mstep}) and \(\widehat{\mLambda}^{+}\) is the corresponding optimal dual variable, under the Karush–Kuhn–Tucker (KKT) conditions \citep{1999Jorge}, we have the following holds,
\begin{equation}\label{A2}
(\mS^l+\mS^u - 2(\widehat{\mTheta}^+)^{-1} +\widehat{\mLambda}^{+})_{ij} + (1/|P|) \dot{p}_\lambda(|\theta^+_{ij}|) = 0, \ \forall i = 1,2,\cdots, nw, j = 1,2,\cdots, nw, \text{and } i \neq j, 
\end{equation}

\begin{equation} \label{A3}
(\mS^l+\mS^u - 2(\widehat{\mTheta}^+)^{-1} +\widehat{\mLambda}^{+})_{ii} = 0, \ \forall i = 1,2,\cdots, nw, 
\end{equation}

\begin{equation}
\widehat{\mTheta}^{+} = \widehat{\mGamma}^{+}, 
\end{equation}

\begin{equation}
\widehat{\mGamma}^+ \in \mathcal{T}, 
\end{equation}
and
\begin{equation}\label{kkt5}
\langle  \widehat{\mLambda}^{+}, \mGamma -\widehat{\mGamma}^{+} \rangle  \leq 0,\ \forall \mGamma \in \mathcal{T}.
\end{equation}

The optimality condition of the subproblem (\ref{mGammastep}) with respect to \(\mGamma\) is given by
\begin{equation} \label{optimalgamma}
\langle \mLambda^{(q)} - (1/\rho)(\mTheta^{(q)} - \mGamma^{(q+1)}), \mGamma-  \mGamma^{(q+1)}\rangle \leq 0, \ \forall \mGamma \in \mathcal{T}.
\end{equation}

Using the update formula of \(\mLambda\), that is, 
\begin{equation}\label{A8}
\mLambda^{(q+1)} = \mLambda^{(q)} - (1/\rho)(\mTheta^{(q+1)} -\mGamma^{(q+1)}), 
\end{equation}
equation (\ref{optimalgamma}) can be rewritten as
\begin{equation}\label{a7rewritten}
\langle \mLambda^{(q+1)} + (1/\rho)(\mTheta^{(q+1)} -\mTheta^{(q)} ), \mGamma-  \mGamma^{(q+1)}\rangle \leq 0, \ \forall \mGamma \in \mathcal{T}.
\end{equation} 

Since equations (\ref{kkt5}) and (\ref{a7rewritten}) are hold for any \(\mGamma \in \mathcal{T}\), replacing \(\mGamma\) with \(\mGamma^{(q+1)}\) in (\ref{kkt5}) and replacing \(\mGamma\) with \(\widehat{\mGamma}^+\) in (\ref{a7rewritten}) yields
\begin{equation} \label{A10}
\langle  \widehat{\mLambda}^{+}, \mGamma^{(q+1)} -\widehat{\mGamma}^{+} \rangle  \leq 0,
\end{equation}
and
\begin{equation} \label{A11}
\langle \mLambda^{(q+1)} +  (1/\rho)(\mTheta^{(q+1)} -\mTheta^{(q)} ), \widehat{\mGamma}^+ -  \mGamma^{(q+1)}\rangle \leq 0.
\end{equation}

From the equations (\ref{A10}) and (\ref{A11}), we have
\begin{equation}\label{A12}
\langle \mGamma^{(q+1)} - \widehat{\mGamma}^+, (\mLambda^{(q+1)} - \widehat{\mLambda}^{+}) + (1/\rho)(\mTheta^{(q+1)} -\mTheta^{(q)} )\rangle \geq 0.
\end{equation}

The optimality condition of the subproblem (\ref{mThetastep}) with respect to \(\mTheta\) is given by
\begin{equation}\label{A13}
\begin{aligned}
0 \in (\mS^l + \mS^u -2(\mTheta^{(q+1)})^{-1})_{ij} &+ (1/|P|) \dot{p}_\lambda(|\theta^{(q+1)}_{ij}|) + \mLambda^{(q)}_{ij} + (1/\rho)(\mTheta^{(q+1)} - \mGamma^{(q+1)})_{ij}, \\
&\forall i = 1,2,\cdots, nw, j = 1,2,\cdots, nw, \text{and } i \neq j, 
\end{aligned}
\end{equation}
and
\begin{equation}\label{A14}
(\mS^l + \mS^u -2(\mTheta^{(q+1)})^{-1})_{ii} + \mLambda^{(q)}_{ii} + (1/\rho)(\mTheta^{(q+1)} - \mGamma^{(q+1)})_{ii} = 0,\ \forall i = 1,2,\cdots, nw.
\end{equation}

Using the update formula of \(\mLambda\), equations (\ref{A8}), (\ref{A13}) and (\ref{A14}), we have
\begin{equation}\label{A15}
\begin{aligned}
(-\mLambda^{(q+1)} -\mS^l - \mS^u + 2(\mTheta^{(q+1)})^{-1})_{ij} &\in (1/|P|)  \dot{p}_\lambda(|\theta^{(q+1)}_{ij}|), \\
& \forall i = 1,2,\cdots, nw, j = 1,2,\cdots, nw, \text{and } i \neq j, 
\end{aligned}
\end{equation}
and
\begin{equation}\label{A16}
(\mS^l + \mS^u - 2(\mTheta^{(q+1)})^{-1})_{ii} + \mLambda^{(q+1)}_{ii} = 0,  \forall i = 1,2,\cdots, nw.
\end{equation}

Using the fact that \(\dot{p}_\lambda(\cdot)\) is a monotonic function, summarizing (\ref{A2}), (\ref{A3}), (\ref{A15}) and (\ref{A16}) yields
\begin{equation}\label{A17}
\langle \mTheta^{(q+1)} - \widehat{\mTheta}^+,  (\widehat{\mLambda}^+ - \mLambda^{(q+1)}) + 2((\mTheta^{(q+1)})^{-1} - (\widehat{\mTheta}^{+})^{-1})\rangle \geq 0.
\end{equation}

From eqautions (\ref{A12}) and (\ref{A17}), we have
\begin{equation} \label{A18}
\begin{aligned}
\lVert \mTheta^{(q+1)} - \widehat{\mTheta}^+  \rVert^2_F &\leq (1/2) \langle \mTheta^{(q+1)} - \widehat{\mTheta}^+,  (\widehat{\mLambda}^+ - \mLambda^{(q+1)})\rangle \\
& + \langle \mGamma^{(q+1)} - \widehat{\mGamma}^+, (\mLambda^{(q+1)} - \widehat{\mLambda}^{+})\rangle \\
&+ \langle \mGamma^{(q+1)} - \widehat{\mGamma}^+, (1/\rho)(\mTheta^{(q+1)} -\mTheta^{(q)} )\rangle, 
\end{aligned}
\end{equation}
using the update formula of \(\mLambda\) in, that is, 
\[
\mLambda^{(q+1)} = \mLambda^{(q)} - (1/\rho)(\mTheta^{(q+1)} -\mGamma^{(q+1)}), 
\]
and \(\widehat{\mGamma}^{+} = \widehat{\mTheta}^{+}\), equations (\ref{A18}) can be rewritten as
\begin{equation} \label{A19}
\begin{aligned}
\lVert \mTheta^{(q+1)} - \widehat{\mTheta}^+  \rVert^2_F &\leq (1/2) \langle \mTheta^{(q+1)} - \widehat{\mTheta}^+,  (\widehat{\mLambda}^+ - \mLambda^{(q+1)})\rangle \\
& + \langle \rho(\mLambda^{(q+1)} - \mLambda^{(q)}) + \mTheta^{(q+1)} - \widehat{\mTheta}^+, (\mLambda^{(q+1)} - \widehat{\mLambda}^{+})\rangle \\
&+ \langle \rho(\mLambda^{(q+1)} - \mLambda^{(q)}) + \mTheta^{(q+1)} - \widehat{\mTheta}^+, (1/\rho)(\mTheta^{(q+1)} -\mTheta^{(q)} )\rangle.
\end{aligned} 
\end{equation}

With some simple arithmetic derivation of (\ref{A19}) leads to
\begin{equation} \label{A20}
\begin{aligned}
\lVert \mTheta^{(q+1)} - \widehat{\mTheta}^+  \rVert^2_F &- \langle\mLambda^{(q+1)} - \mLambda^{(q)}, \mTheta^{(q+1)} -\mTheta^{(q)}\rangle \\
& \leq (1/2) \langle \mTheta^{(q+1)} - \widehat{\mTheta}^+,  (\mLambda^{(q+1)} - \widehat{\mLambda}^+)\rangle  + \rho\langle (\mLambda^{(q+1)} - \mLambda^{(q)}), (\mLambda^{(q+1)} - \widehat{\mLambda}^{+})\rangle \\
&+ (1/\rho) \langle \mTheta^{(q+1)} - \widehat{\mTheta}^+, (\mTheta^{(q+1)} -\mTheta^{(q)} )\rangle\\
& \leq \rho \langle (\mLambda^{(q+1)} - \mLambda^{(q)}), (\mLambda^{(q+1)} - \widehat{\mLambda}^{+})\rangle 
+ (1/\rho) \langle \mTheta^{(q+1)} - \widehat{\mTheta}^+, (\mTheta^{(q+1)} -\mTheta^{(q)} )\rangle, \\
\end{aligned}
\end{equation}
with
\[
\begin{aligned}
\widehat{\mLambda}^{+} - \mLambda^{(q+1)} & = (\widehat{\mLambda}^{+} - \mLambda^{(q)}) +(\mLambda^{(q)} - \mLambda^{(q+1)}), \\
\widehat{\mTheta}^{+} - \mTheta^{(q+1)} & = (\widehat{\mTheta}^{+} - \mTheta^{(q)}) +(\mTheta^{(q)} - \mTheta^{(q+1)}), 
\end{aligned}
\]
equation (\ref{A20}) can be reduced to
\begin{equation} \label{A21}
\begin{aligned}
& \rho \langle \mLambda^{(q)} - \widehat{\mLambda}^{+}, \mLambda^{(q)}- \mLambda^{(q+1)}\rangle + \rho \langle \mTheta^{(q)} - \widehat{\mTheta}^{+}, \mTheta^{(q)}- \mTheta^{(q+1)}\rangle\\
& \geq \rho \lVert \mLambda^{(q)}- \mLambda^{(q+1)} \rVert^2_F + (1/\rho) \lVert\mTheta^{(q)}- \mTheta^{(q+1)} \rVert^2_F \\
&  + \lVert \mTheta^{(q+1)} - \widehat{\mTheta}^+  \rVert^2_F - \langle\mLambda^{(q+1)} - \mLambda^{(q)}, \mTheta^{(q+1)} -\mTheta^{(q)}\rangle \\
\end{aligned}
\end{equation}

Recall that
\[
\mD = \left[
\begin{array}{cc}
	\rho \mI_{nw\times nw} & \bm{0}\\
	\bm{0} & (1/\rho) \mI_{nw\times nw} 
\end{array}
\right],
\]
using the notation of \(\mU^{(q)}\) and \(\mU^+\), equation (\ref{A21}) can be reduced to
\begin{equation} \label{A22}
\langle \mU^{(q)} - \mU^{+},   \mU^{(q)}- \mU^{(q+1)}\rangle_\mD \geq  \lVert \mU^{(q)} - \mU^{(q+1)} \rVert^2_\mD + \lVert \mTheta^{(q+1)} - \widehat{\mTheta}^+  \rVert^2_F - \langle\mLambda^{(q+1)} - \mLambda^{(q)}, \mTheta^{(q+1)} -\mTheta^{(q)}\rangle.  \\
\end{equation}

According to the definition of \(\langle \cdot, \cdot\rangle_{\mD}\), we have following identity
\[
\lVert \mU^{(q+1)} - \mU^{+} \rVert^2_\mD =  \lVert \mU^{(q+1)} - \mU^{(q)} \rVert^2_\mD + \lVert \mU^{(q)} - \mU^{+} \rVert^2_\mD- 2 \langle \mU^{(q)} - \mU^{(q+1)}, \mU^{(q)} - \mU^{+} \rangle_\mD,
\]
combing with equation (\ref{A22}), we get,
\begin{equation}\label{A23}
\begin{aligned}
& \lVert \mU^{(q)} - \mU^{+} \rVert^2_\mD - \lVert \mU^{(q+1)} - \mU^{+} \rVert^2_\mD = 2 \langle \mU^{(q)} - \mU^{(q+1)}, \mU^{(q)} - \mU^{+} \rangle_\mD - \lVert \mU^{(q+1)} - \mU^{(q)} \rVert^2_\mD \\
& \geq 2 \lVert \mU^{(q)} - \mU^{(q+1)} \rVert^2_\mD + 2\lVert \mTheta^{(q+1)} - \widehat{\mTheta}^+  \rVert^2_F  - 2\langle\mLambda^{(q+1)} - \mLambda^{(q)}, \mTheta^{(q+1)} -\mTheta^{(q)}\rangle - \lVert \mU^{(q+1)} - \mU^{(q)} \rVert^2_\mD\\
& = \lVert \mU^{(q)} - \mU^{(q+1)} \rVert^2_\mD + 2\lVert \mTheta^{(q+1)} - \widehat{\mTheta}^+  \rVert^2_F - 2\langle\mLambda^{(q+1)} - \mLambda^{(q)}, \mTheta^{(q+1)} -\mTheta^{(q)}\rangle.
\end{aligned}
\end{equation}

According to equations (\ref{A15}) and (\ref{A16}), we have
\begin{equation}\label{A24}
	\begin{aligned}
		(-\mLambda^{(q)} -\mS^l - \mS^u + 2(\mTheta^{(q)})^{-1})_{ij} &\in (1/|P|)  \dot{p}_\lambda(|\theta^{(q)}_{ij}|), \\
		& \forall i = 1,2,\cdots, nw, \ j = 1,2,\cdots, nw, \text{and } i \neq j, 
	\end{aligned}
\end{equation}
and
\begin{equation}\label{A25}
	(\mS^l + \mS^u - 2(\mTheta^{(q)})^{-1})_{ii} + \mLambda^{(q)}_{ii} = 0,  \forall i = 1,2,\cdots, nw.
\end{equation}

Similarly, using the fact that \(\dot{p}_\lambda(\cdot)\) is a monotonic function and combing equations (\ref{A24}) and (\ref{A25}), we have
\[
- \langle\mLambda^{(q+1)} - \mLambda^{(q)}, \mTheta^{(q+1)} -\mTheta^{(q)}\rangle \geq \lVert \mTheta^{(q+1)} -\mTheta^{(q)} \rVert^2_F \geq 0,
\]
then equation (\ref{A23}) reduced to
\[
\lVert \mU^{(q)} - \mU^{+} \rVert^2_\mD - \lVert \mU^{(q+1)} - \mU^{+} \rVert^2_\mD \geq \lVert \mU^{(q)} - \mU^{(q+1)} \rVert^2_\mD,
\]
which completes the proof.
\end{proof}

{\bf Theorem 4.1} {\it 
The sequence \(\left\{\left(\mTheta^{(q)}, \mGamma^{(q)}, \mLambda^{(q)}\right)\right\}\) produced by ADMM algorithm from any starting point converges to an optimal solution of (\ref{Mstep}), that is, 

(a) \(\lVert \mU^{(q)} - \mU^{(q+1)}\rVert_\mD \to 0\);

(b) \(\left\{\mU^{(q)}\right\}\) located in a compact region;

(c) \(\lVert \mU^{(q)} - \mU^{+}\rVert_\mD\) is monotonically non-increasing.
}

\begin{proof}
Using Lemma \ref{lemma41}, that is, 
\[
\lVert \mU^{(q)} - \mU^{+} \rVert^2_\mD - \lVert \mU^{(q+1)} - \mU^{+} \rVert^2_\mD \geq \lVert \mU^{(q)} - \mU^{(q+1)} \rVert^2_\mD
\]
and \(\lVert \mU^{(q)} - \mU^{(q+1)} \rVert^2_\mD \geq 0\), we have 
\[
\lVert \mU^{(q)} - \mU^{+} \rVert^2_\mD \geq \lVert \mU^{(q+1)} - \mU^{+} \rVert^2_\mD.
\]

Therefore, the conclusion (a), (b) and (c) can be obtained easily. It follows from (a) that \(\lVert \mLambda^{(q)} - \mLambda^{(q+1)}\rVert^2_F \to 0\) and \(\lVert \mTheta^{(q)} -\mTheta^{(q+1)}\rVert^2_F \to 0\), that is, \(\mLambda^{(q)} \to \mLambda^{(q+1)}\) and \( \mTheta^{(q)} \to \mTheta^{(q+1)}\). Using the update formula of \(\mLambda\), that is, 
\[
\mLambda^{(q+1)} = \mLambda^{(q)} - (1/\rho)(\mTheta^{(q+1)} -\mGamma^{(q+1)}), 
\]
we have \(\mTheta^{(q+1)} \to \mGamma^{(q+1)}\). From (b) it can be obtained that there exists a subsequence \(\left\{\mU^{i_j}\right\}\) of \(\left\{\mU\right\}\) converging to \(\bar{\mU} = (\bar{\mTheta}, \bar{\mLambda})\), that is, \(\mLambda^{i_j} \to \bar{\mLambda}\) and \( \mTheta^{i_j} \to \bar{\mTheta}\). Simultaneously, we can get \(\mGamma^{i_j}\to \bar{\mGamma} := \bar{\mTheta}\). The above discussion illustrates that \((\bar{\mTheta}, \bar{\mLambda}, \bar{\mGamma})\) is a limit point of \(\left\{(\bar{\mTheta}^{(q)}, \bar{\mLambda}^{(q)}, \bar{\mGamma}^{(q)})\right\}\). 

According to equations (\ref{A15}) and (\ref{A16}), we have
\begin{equation}\label{A26}
	\begin{aligned}
		(-\bar{\mLambda} -\mS^l - \mS^u + 2(\bar{\mTheta})^{-1})_{ij} &\in (1/|P|)  \dot{p}_\lambda(|\bar{\theta}_{ij}|), \\
		& \forall i = 1,2,\cdots, nw, j = 1,2,\cdots, nw, \text{and } i \neq j, 
	\end{aligned}
\end{equation}
and
\begin{equation}\label{A27}
	(\mS^l + \mS^u - 2(\bar{\mTheta})^{-1})_{ii} + \bar{\mLambda}_{ii} = 0,  \forall i = 1,2,\cdots, nw,  
\end{equation}
and combing with equation (\ref{a7rewritten})
\begin{equation}\label{A28}
\langle \bar{\mLambda}, \mGamma- \bar{\mGamma}\rangle \leq 0, \ \forall \mGamma \in \mathcal{T}.
\end{equation}

Equations (\ref{A26}) and (\ref{A27}) together with equation (\ref{A28}) implies that \((\bar{\mTheta}, \bar{\mLambda}, \bar{\mGamma})\) is an optimal solution to (\ref{Mstepfisrststage}), which completes the proof.
\end{proof}

\section{The computation procedure of equation (\ref{partialdevwork})} \label{mThetastepcompapp}
We first decompose the matrices \(\mW\), \(\mG^{(q)}\), and \(\mTheta\) into the following block forms:
\[
\mW =\left[
\begin{array}{cc}
	\mW_{11} & \vw_{12}\\
	\vw_{21} &w_{22}\\ 
\end{array}
\right], \ \mG =\left[
\begin{array}{cc}
	\mG^{(q)}_{11} & \vg^{(q)}_{12}\\
	\vg^{(q)}_{21} &g^{(q)}_{22}\\ 
\end{array}
\right],\ \mTheta =\left[
\begin{array}{cc}
	\mTheta_{11} & \vtheta_{12}\\
	\vtheta_{21} &\theta_{22}\\ 
\end{array}
\right],
\]
where \(\mW_{11}, \mG^{(q)}_{11}, \mTheta_{11} \in \mathbb{R}^{(nw -1) \times (nw -1)}\), \(\vw_{12} , \vg^{(q)}_{12}, \vtheta_{12} \in \mathbb{R}^{(nw -1)}\), \(w_{22}, g^{(q)}_{22}, \theta_{22} \in \mathbb{R}\), and 
\[
\vw_{21} = \vw^\top_{12},\  \vg^{(q)}_{21} = (\vg^{(q)}_{12})^\top,\ \vtheta_{21} = \vtheta^ \top_{12}.
\]

Based on the fundamental formula for block matrix inversion, we have
\[
\mW =\left[
\begin{array}{cc}
	\mW_{11} & \vw_{12}\\
	\vw_{21} &w_{22}\\ 
\end{array}
\right] = \left[
\begin{array}{cc}
	(\mTheta_{11} - (\vtheta_{12}\vtheta_{21}) / \theta_{22})^{-1} &  -\mW_{11} \vtheta_{12} / \theta_{22}\\
	-\vtheta_{21}(\mTheta_{11} - \vtheta_{12} \theta^{-1}_{22} \vtheta_{21})^{-1}/\theta_{22} & (\theta_{22} + \vtheta_{21} \mW_{11} \vtheta_{12}) / \theta^2_{22}
\end{array}
\right].
\]

Partition \(\mS^l\), \(\mS^u\), \(\mLambda^{(q)}\), \(\mGamma^{(q+1)}\), and \(\mA\) into the same block form as \(\mW\),
\[
\mS^l =\left[
\begin{array}{cc}
	\mS^l_{11} & \vs^l_{12}\\
	\vs^l_{21} &s^l_{22}\\ 
\end{array}
\right], 
\mS^u =\left[
\begin{array}{cc}
	\mS^u_{11} & \vs^u_{12}\\
	\vs^u_{21} &s^u_{22}\\ 
\end{array}
\right],
\mLambda^{(q)} =\left[
\begin{array}{cc}
	\mLambda^{(q)}_{11} & \vlambda^{(q)}_{12}\\
	\vlambda^{(q)}_{21} &\lambda^{(q)}_{22}\\ 
\end{array}
\right],
\]

\[
\mGamma^{(q+1)}=\left[
\begin{array}{cc}
	\mGamma^{(q+1)}_{11} & \vgamma^{(q+1)}_{12}\\
	\vgamma^{(q+1)}_{21} & \gamma^{(q+1)}_{22}\\ 
\end{array}
\right],
\mA =\left[
\begin{array}{cc}
	\mA_{11} & \va_{12}\\
	\va_{21} &a_{22}\\ 
\end{array}
\right].
\]

Without loss of generality, consider the \(nw\)-th column of equation (\ref{partialdevwork}) without diagonal elements, which corresponds to the following equality
\[ 
\vs^{l}_{12} + \vs^{u}_{12}- 2\vw_{12} + \vg^{(q)}_{12}\odot \va_{12} + \vlambda^{(q)}_{12} +(1/\rho)(\vtheta_{12} - \vgamma^{(q+1)}_{12}) = \bm{0}.
\]

Let \(\vbeta = (\beta_1, \beta_2, \cdots, \beta_{nw-1})^\top\coloneqq - \vtheta_{12} / \theta_{22} \in \mathbb{R}^{nw-1}\). Using the fact that \(\vw_{12} = -\mW_{11} \vtheta_{12} / \theta_{22}\), we have
\begin{equation} \label{factbeta}
	2\mW_{11}\vbeta + (\theta_{22}/\rho)\vbeta - \vg^{(q)}_{12}\odot \va_{12} -\left(\vs^{l}_{12} + \vs^{u}_{12} + \vlambda^{(q)}_{12} - (1/\rho)\vgamma^{(q+1)}_{12}\right) = \bm{0}. 
\end{equation}

It is evident that (\ref{factbeta}) corresponds to the normal equations of the following quadratic programming problem
\[
\min_{\vbeta} \vbeta^\top \mW_{11} \vbeta - \vbeta^\top\left(\vs^{l}_{12} + \vs^{u}_{12} + \vlambda^{(q)}_{12} - (1/\rho)\vgamma^{(q+1)}_{12}\right)+ (\theta_{22}/2\rho)\lVert \vbeta \rVert^2_2- \sum_{i\leq nw-1} (\vg^{(q)}_{12})_i|\beta_{i}|
\]
or equivalently
\begin{equation} \label{eququadratipro}
	\min_\vbeta \bigg \lVert\mW^{1/2}_{11}\vbeta -\mW^{-1/2}_{11}\left(\vs^{l}_{12} + \vs^{u}_{12} + \vlambda^{(q)}_{12} - (1/\rho)\vgamma^{(q+1)}_{12}\right) \bigg \rVert^2_2 + (\theta_{22}/2\rho)\lVert \vbeta \rVert^2_2 -\sum_{i\leq nw-1} (\vg^{(q)}_{12})_i|\beta_{i}|.
\end{equation}

The optimization objective (\ref{eququadratipro}) in the quadratic programming problem can be transformed into the following standard elastic net form
\[
Q(\vbeta) =  \lVert \vx - \mZ\vbeta \rVert^2_2 + (\theta_{22}/2\rho)\lVert \vbeta \rVert^2_2 + \sum_{i\leq nw-1} (\vg^{(q)}_{12})_i|\beta_{i}|, 
\]
where
\[
\mZ =\mW^{1/2}_{11}, \vx = \mW^{-1/2}_{11}\left(\vs^{l}_{12} + \vs^{u}_{12} + \vlambda^{(q)}_{12} - (1/\rho)\vgamma^{(q+1)}_{12}\right).
\]

Setting the subgradient of \(Q(\vbeta)\) with respect to \(\vbeta\) to zero, we obtain the estimate of \(\vbeta\) as
\begin{equation}\label{Thetabeta}
	\widehat{\beta}_j = \frac{S_{j}\left(\left(\vs^{l}_{12} + \vs^{u}_{12} + \vlambda^{(q)}_{12} - (1/\rho)\vgamma^{(q+1)}_{12}\right)_j- \sum_{i \neq j}(\mW_{11})_{ij} \widehat{\beta}_i\right) }{(\mW_{11})_{jj} +(\theta_{22}/2\rho)}, \ j = 1,2,\cdots, nw-1,
\end{equation}
where \(S_j(\cdot) = \sign(\cdot)(|\cdot| -(\vg^{(q)}_{12})_j)\) is the soft-thresholding operator. Based on the estimated \(\widehat{\vbeta} = (\widehat{\beta}_1, \widehat{\beta}_2,\cdots, \widehat{\beta}_{nw-1})^{\top}\), we update the other entries according to the following three steps: 
\begin{enumerate}
	\item[(\uppercase\expandafter{\romannumeral1})] \(\widehat{\vw}_{12} = -\mW_{11}\vtheta_{12} / \theta_{22} = \mW_{11} \widehat{\vbeta}\),
	\item[(\uppercase\expandafter{\romannumeral2})] \(\widehat{\theta}_{22} = 1/(w_{22} - \widehat{\vbeta}\widehat{\vw}_{12})\), \(\widehat{\vtheta}_{12} = -\widehat{\theta}_{22}\widehat{\vbeta}\),
	\item[(\uppercase\expandafter{\romannumeral3})] \(\widehat{w}_{22} = (1/2)(s^l_{22} + s^u_{22} + \lambda^{(q)}_{22} + (1/\rho)(\widehat{\theta}_{22} - \gamma^{(q+1)}_{22}))\).
\end{enumerate}

Algorithm \ref{completeMstep} shows the complete details of the parameters update procedure of (\ref{Mstep}). 
\begin{algorithm}[H]
\caption{Parameters update procedure of (\ref{Mstep})}
\label{completeMstep}
\begin{algorithmic}
\REQUIRE ~~\\
Input \(\mTheta^{(q)}, \mLambda^{(q)}, \rho\).
\ENSURE ~~\\	
For the \((q+1)\)-th iteration\\
1. Solve \(\mGamma^{(q+1)}\) based on (\ref{Tclosr});\\
2. Initialize \(\mW = (\mTheta^{(q)})^{-1}\);\\
3. Solve \(\mTheta^{(q+1)}\) by cycling around the columns repeatedly, that is,\\
\qquad 3.1 Solve \(\widehat{\vbeta}\) based on (\ref{Thetabeta}),\\
\qquad 3.2 Solve \(\widehat{\vw}_{12} = \mW_{11} \widehat{\vbeta}\),\\
\qquad 3.3 Solve \(\widehat{w}_{22} = (1/2)(s^l_{22} + s^u_{22} + \lambda^{(q)}_{22} + (1/\rho)(\widehat{\theta}_{22} - \gamma^{(q+1)}_{22}))\),\\
4. Repeat the above cycle 3.1-3.3 till convergence;\\
5. Repeat steps 1-4 till convergence.\\
\textbf{Output}: Block Toeplitz sparse precision matrix \(\widehat{\mTheta}\).\\
\end{algorithmic}
\end{algorithm}

\newpage
\section{Table of selected stocks}\label{informationsselected}
\begin{table}[H]
\setlength{\abovecaptionskip}{0pt}
\setlength{\belowcaptionskip}{3pt}
\centering
\caption{The selected stock abbreviations and their corresponding sectors.}
\label{selectedstocktable1}
\setlength{\tabcolsep}{1.2mm}{ 
\begin{tabular}{cll}
\toprule
\textbf{Sector} & \textbf{Stock symbol} & \textbf{Company}\\
\hline
\multirow{8}{*}{\tabincell{c}{Basic Materials (8)}} & \$XOM  & Exxon Mobil Corporation\\
& \$RDS-B & Royal Dutch Shell plc\\
& \$CVX & Chevron Corporation\\
& \$TOT & TOTAL S.A. \\
& \$BP & BP p.l.c.\\
& \$BHP & BHP Billiton Limited\\
& \$SLB & Schlumberger Limited\\
& \$BBL & BHP Billiton plc\\
\hline

\multirow{10}{*}{\tabincell{c}{Consumer Goods (10)}} & \$AAPL & Apple Inc.\\
& \$PG & The Procter \& Gamble Company\\
& \$BUD & Anheuser-Busch InBev SA/NV \\
& \$KO & The Coca-Cola Company\\
& \$PM & Philip Morris International Inc.\\
& \$TM & Toyota Motor Corporation \\
& \$PEP & Pepsico, Inc. \\
& \$UN & Unilever N.V. \\
& \$UL & Unilever PLC \\
& \$MO & Altria Group, Inc. \\
\hline

\multirow{9}{*}{\tabincell{c}{Healthcare (9)}} & \$JNJ & Johnson \& Johnson\\
& \$PFE & Pfizer Inc. \\
& \$NVS & Novartis AG \\
& \$UNH & UnitedHealth Group Incorporated \\
& \$MRK & Merck \& Co., Inc. \\
& \$AMGN & Amgen Inc.\\
& \$MDT & Medtronic plc\\
& \$SNY & Sanofi\\
& \$CELG & Celgene Corporation\\
\hline

\multirow{9}{*}{\tabincell{c}{Services (9)}} & \$AMZN  & Amazon.com, Inc.\\
& \$WMT & Wal-Mart Stores, Inc.\\
& \$CMCSA & Comcast Corporation \\
& \$HD & The Home Depot, Inc.\\
& \$DIS & The Walt Disney Company \\
& \$MCD & McDonald’s Corporation\\
& \$CHTR & Charter Communications, Inc.\\
& \$UPS & United Parcel Service, Inc.\\
& \$PCLN & The Priceline Group Inc.\\ 
\bottomrule
\end{tabular}
}
\end{table}

\begin{table}[H]
\setlength{\abovecaptionskip}{0pt}
\setlength{\belowcaptionskip}{3pt}
\centering
\caption{The selected stock abbreviations and their corresponding sectors (Continued Table \ref{selectedstocktable1}).}
\label{selectedstocktable2}
\setlength{\tabcolsep}{1.2mm}{ 
\begin{tabular}{cll}
\toprule
\textbf{Sector} & \textbf{Stock symbol} & \textbf{Company}\\
\hline
\multirow{10}{*}{\tabincell{c}{Utilities (10)}} & \$NEE &  NextEra Energy, Inc. \\
&\$DUK & Duke Energy Corporation \\
&\$D & Dominion Energy, Inc. \\
&\$SO & The Southern Company  \\
&\$NGG&  National Grid plc \\
&\$AEP &American Electric Power Company, Inc. \\
&\$PCG& PG\&E Corporation \\
&\$EXC &Exelon Corporation  \\
&\$SRE& Sempra Energy \\
&\$PPL &PPL Corporation \\
\hline

\multirow{5}{*}{\tabincell{c}{Conglomerates (5)}} & \$IEP & Icahn Enterprises L.P.\\
&  \$HRG & HRG Group, Inc. \\
& \$CODI & Compass Diversified Holdings LLC \\
& \$SPLP & Steel Partners Holdings L.P. \\
& \$PICO & PICO Holdings, Inc. \\
\hline

\multirow{10}{*}{\tabincell{c}{Financial (10)}} & \$BCH & Banco de Chile \\
& \$BSAC & Banco Santander-Chile \\
& \$BRK-A &  Berkshire Hathaway Inc. \\
& \$JPM & JPMorgan Chase \& Co. \\
& \$WFC & Wells Fargo \& Company \\
& \$BAC & Bank of America Corporation \\
& \$V & Visa Inc. \\
& \$C & Citigroup Inc. \\
& \$HSBC &  HSBC Holdings plc \\
& \$MA & Mastercard Incorporated\\ 
\hline

\multirow{10}{*}{\tabincell{c}{Industrial Goods (10)}} & \$GE & General Electric Company \\
& \$MMM & 3M Company\\
& \$BA & The Boeing Company \\
& \$HON & Honeywell International Inc. \\
& \$UTX & United Technologies Corporation \\
& \$LMT & Lockheed Martin Corporation\\
& \$CAT &  Caterpillar Inc.\\
& \$GD & General Dynamics Corporation \\
& \$DHR & Danaher Corporation \\
& \$ABB & ABB Ltd\\
\bottomrule
\end{tabular}
}
\end{table}

\begin{table}[H]
\setlength{\abovecaptionskip}{0pt}
\setlength{\belowcaptionskip}{3pt}
\centering
\caption{The selected stock abbreviations and their corresponding sectors (Continued Table \ref{selectedstocktable2}).}
\label{selectedstocktable3}
\setlength{\tabcolsep}{1.2mm}{
\begin{tabular}{cll} 
\toprule
\textbf{Sector} & \textbf{Stock symbol} & \textbf{Company}\\
\hline
\multirow{10}{*}{\tabincell{c}{Technology (10)}} & \$GOOG  & Alphabet Inc.\\
& \$MSFT & Microsoft Corporation \\
& \$FB & Facebook, Inc. \\
& \$T & AT\&T Inc. \\
& \$CHL &  China Mobile Limited \\
& \$ORCL & Oracle Corporation  \\
& \$TSM & Taiwan Semiconductor Manufacturing Company Limited  \\
& \$VZ & Verizon Communications Inc. \\
& \$INTC & Intel Corporation \\ 
& \$CSCO&  Cisco Systems, Inc. \\
\bottomrule
\end{tabular}
}
\end{table}

\end{appendix}

\newpage
\bibliography{./bib/bib.bib}
\bibliographystyle{elsarticle-num-names}

\end{document}